\documentclass[10pt,twocolumn,letterpaper]{article}

\usepackage[T1]{fontenc}
\usepackage[utf8]{inputenc}
\usepackage[margin=0.75in]{geometry}
\usepackage{times}
\usepackage{titlesec}
\titlespacing*{\subsection}{0pt}{6pt plus 1pt minus 1pt}{3pt plus 1pt minus 1pt}
\usepackage{xspace}
\usepackage[dvipsnames]{xcolor}
\usepackage{graphicx,amssymb,booktabs}
\usepackage[numbers,sort&compress]{natbib}
\usepackage[hyphens]{url}
\usepackage[format=plain,labelformat=simple,labelsep=period,font=small]{caption}
\usepackage[font=footnotesize,skip=3pt,subrefformat=parens]{subcaption}
\usepackage[shortlabels,inline]{enumitem}

\usepackage{amsmath,amsfonts,bm}

\def\eqref#1{equation~\ref{#1}}

\def\1{\bm{1}}

\DeclareMathAlphabet{\mathsfit}{\encodingdefault}{\sfdefault}{m}{sl}
\SetMathAlphabet{\mathsfit}{bold}{\encodingdefault}{\sfdefault}{bx}{n}

\usepackage{placeins,float,needspace,pdflscape}
\usepackage{tikz}
\usetikzlibrary{positioning,arrows.meta}
\graphicspath{{figures/}}
\definecolor{linkblue}{rgb}{0.21,0.49,0.74}
\usepackage[breaklinks,colorlinks,allcolors=linkblue]{hyperref}

\title{Don’t Handcraft, Learn: Generalizable Monitor Functions for Adaptive Mesh Movement via Variational Information Bottleneck}
\author{%
  \begin{tabular}{ccc}
    \shortstack{Zhixiang Wang$^{1}$\\{\scriptsize\texttt{22920242203422@stu.xmu.edu.cn}}} &
    \shortstack{Can Yang$^{1}$\\{\scriptsize\texttt{yangcan@stu.xmu.edu.cn}}} &
    \shortstack{Yunru Chen$^{2}$\\{\scriptsize\texttt{112052024018@student.fjnu.edu.cn}}} \\[0.8em]
    \multicolumn{3}{c}{%
      \begin{tabular}{cc}
        \shortstack{Zhenzhong Wang$^{1}$\thanks{Corresponding author: \texttt{zhenzhongwang@xmu.edu.cn}}\\{\scriptsize\texttt{zhenzhongwang@xmu.edu.cn}}} &
        \shortstack{Min Jiang$^{1}$\\{\scriptsize\texttt{minjiang@xmu.edu.cn}}}
      \end{tabular}}
  \end{tabular}\\[0.8em]
  {\small $^{1}$Xiamen University\qquad $^{2}$Fujian Normal University}
}
\date{}
\makeatletter
\let\phyibpapertitle\@title
\makeatother
\hypersetup{
  pdftitle={Don’t Handcraft, Learn: Generalizable Monitor Functions for Adaptive Mesh Movement via Variational Information Bottleneck},
  pdfauthor={Zhixiang Wang, Can Yang, Yunru Chen, Zhenzhong Wang, Min Jiang}
}

\begin{document}
\raggedbottom
\clubpenalty=10000
\widowpenalty=10000

\setcounter{topnumber}{2}
\setcounter{dbltopnumber}{2}
\setcounter{totalnumber}{4}
\renewcommand{\topfraction}{0.85}
\renewcommand{\dbltopfraction}{0.92}
\renewcommand{\bottomfraction}{0.70}
\renewcommand{\textfraction}{0.10}

\makeatletter
\let\phyib@maketitle\@maketitle
\def\@maketitle{%
  \phyib@maketitle
  \begin{center}
\includegraphics[width=0.98\textwidth]{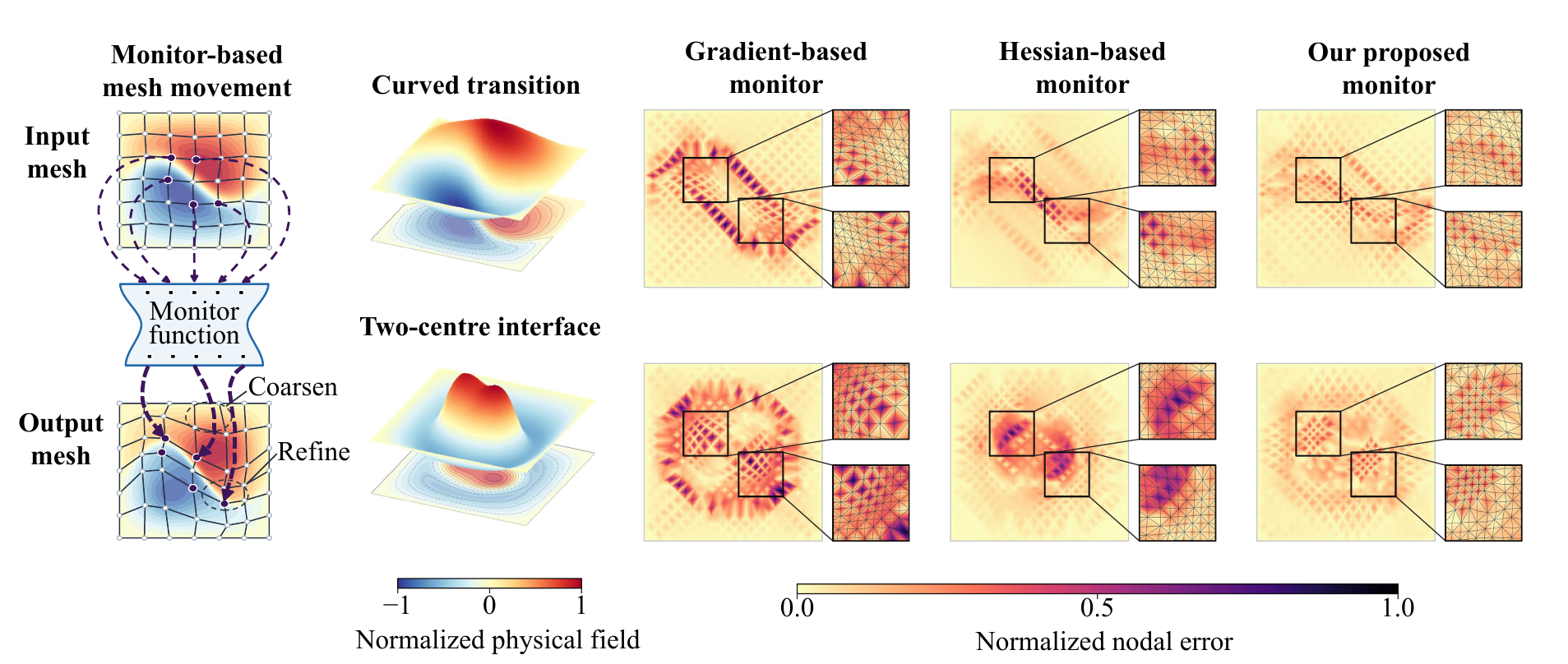}\par
    \vspace{0.25em}
    \begin{minipage}{0.99\textwidth}
      \footnotesize  \refstepcounter{figure}\label{fig:teaser}%
      \textbf{Figure~\thefigure.} The monitor function assigns a monitor value to each mesh node, indicating whether that node should be refined or coarsened. Traditional methods rely on a predefined, fixed monitor function and therefore follow a fixed pattern or bias, leading to suboptimal performance. In contrast, our method learns the monitor values automatically through an information bottleneck.
    \end{minipage}
  \end{center}
}
\makeatother

\maketitle

\begin{abstract}
Many natural phenomena are governed by partial differential equations (PDEs). Learning-based mesh movement aims to use machine learning to redistribute mesh nodes within the spatial domain, concentrating resolution in critical regions to improve PDE solution accuracy. Despite their efficiency, they rely on handcrafted monitor functions to specify where the mesh should be refined or coarsened. Consequently, the mesh mover can only follow a fixed pattern or bias, failing to adapt to the characteristics of the specific PDE. In this work, we introduce PhyIB-Mesh, an information-bottleneck-based framework that learns node-wise importance scores to serve as monitor function values for adaptive mesh movement. Specifically, a graph encoder jointly encodes physical-field information and mesh geometry, producing discriminative node representations that couple the field with the mesh. An information-bottleneck module assigns node-wise importance scores, assigning larger values to nodes whose representations are most relevant for mesh adaptation. These importance values serve directly as monitor function values and can be seamlessly integrated into mesh movement methods that rely on monitor functions.
Experiments across seven two- and three-dimensional benchmarks show competitive numerical accuracy and generalization.

\end{abstract}

\section{Introduction}

Many natural phenomena, such as climate and fluid flow, are governed by partial differential equations (PDEs). To simulate these phenomena, many numerical methods require a computational mesh to discretize the spatial domain on which the PDEs are defined~\cite{xu2025amr,welschinger2026pde}. Finer meshes improve accuracy but increase computational cost. Consequently, mesh quality significantly affects both the accuracy and computational efficiency of numerical solutions~\cite{huang2005metric}. To balance accuracy and efficiency under limited computational cost, $r$-adaptation, or mesh movement, typically refines the mesh in highly dynamic parts of the system while coarsening quasi-stationary regions, without increasing the mesh size~\cite{budd2009moving}. Classical $r$-adaptation methods typically require repeated monitor-field evaluation and iterative mesh-movement solves at each adaptation step, which dominate their computational cost~\cite{cao1999monitor,huang2011adaptive,loseille2011continuous1}.


To reduce this cost, learning-based mesh movement replaces iterative mesh-movement solves with learned node-displacement mappings~\cite{zhang2020meshingnet,huang2021optimalmesh,pfaff2021meshgraphnets,sanchezgonzalez2020gns,brandstetter2022mp_pde}. M2N~\cite{song2022m2n} proposed the first end-to-end learning-based mesh movement framework, introducing the neural spline model and the graph attention network~\cite{velickovic2018gat}, and taking monitor function values as input. UM2N~\cite{zhang2024um2n} introduced a graph-transformer encoder with a graph attention decoder, enabling movement across PDE types, boundary geometries, and mesh structures. UGM2N~\cite{wang2025ugm2n} instead eliminated the need for pre-adapted mesh data by using unsupervised localized geometric feature learning and enforcing nodal equidistribution in a metric space defined by a matrix-valued monitor function.  DMM~\cite{hu2024dmm} learns a data-free neural mesh adaptor by training through the Monge-Ampère equation~\cite{budd2009moving}, mapping the solution directly to adaptive meshes without optimal mesh supervision.

Despite these advances, most learning-based mesh adaptation methods have a fixed monitor that specifies the regions to be refined or coarsened, and the neural network is trained to relocate nodes accordingly. Typically, the monitor function is handcrafted from local solution features—for example, gradient- or Hessian-based monitors compute an “importance” value for each node, which then determines whether the mesh is coarsened or refined~\cite{loseille2011continuous2,zienkiewicz1992recovery,ainsworth2000error}, as illustrated in Figure~\ref{fig:teaser}. In this process, the monitor is fixed. The learned mover can only follow the refinement pattern implied by this monitor and cannot adaptively adjust it according to the specific PDE. Moreover, different monitor functions introduce different biases, leading to suboptimal accuracy and poor generalization across different PDEs.

To mitigate this issue, we propose PhyIB-Mesh, an information-bottleneck-based framework that learns node-wise importance values to serve as monitor function values, instead of relying on a handcrafted monitor. A graph encoder first summarizes the physical-field variation and the mesh geometry at each node. Then, an information-bottleneck module compresses these node representations into a continuous importance value per node that encourages only the nodes most relevant to mesh adaptation to retain large values.
These importance values serve directly as monitor function values and can be seamlessly integrated into mesh movement methods that rely on monitor function values, so that resolution allocation is determined by the PDE itself rather than by a predefined monitor. Our contributions are as follows.


\begin{itemize}
    \item  We propose PhyIB-Mesh, an information-bottleneck-based framework that learns node-wise importance values to serve as monitor function values for mesh movement. Instead of relying on fixed monitor functions, the monitor values are learned according to the PDE itself, improving the numerical accuracy and generalization across PDEs.

    \item We design a graph encoder that encodes scalar states for physical fields and vector states for mesh geometry. By capturing both field variation and mesh stretching, the encoder yields discriminative node representations, providing the information-bottleneck module with a basis for assessing node importance.

    \item We introduce an information-bottleneck-based monitor, solved via a variational strategy and sparsity regularization, that computes node-wise importance values indicating regions of key physical-field variation where the mesh should be refined.

    \item Experiments across two- and three-dimensional PDE settings demonstrate that PhyIB-Mesh achieves strong numerical accuracy and zero-shot generalization, and that the learned node-wise importance values are physically meaningful.
    
\end{itemize}

\section{Preliminaries}

Mesh adaptation can be divided into \(h\)-adaptation and \(r\)-adaptation. \(h\)-adaptation changes mesh connectivity and node count through local refinement or coarsening~\cite{wu2023lamp,yang2023rlamr,foucart2023drlamr}. However, this requires repeated remeshing~\cite{freymuth2025amber} as well as interpolation between meshes, which can be computationally expensive. \(r\)-adaptation, also called mesh movement, keeps connectivity and node count fixed and relocates nodes, thereby avoiding expensive remeshing. For \(r\)-adaptation, M2N~\cite{song2022m2n} proposed the first end-to-end framework, combining neural splines with a graph attention network to generate new mesh. UM2N~\cite{zhang2024um2n} introduced a graph-transformer encoder for zero-shot generalization across PDE types, boundary geometries, and mesh structures. UGM2N~\cite{wang2025ugm2n} used unsupervised localized geometric feature learning and an M-uniform loss in a metric space defined by a matrix-valued monitor. DMM~\cite{hu2024dmm} learned a data-free adaptor through the Monge--Amp\`ere equation, and G-Adaptivity~\cite{rowbottom2025gadaptivity} trained a GNN to directly minimize finite-element solution error via a differentiable solver.


These learning-based mesh adaptation methods can be formally formulated as follows: Let $\Omega\subset\mathbb{R}^d$, $d\in\{2,3\}$, and let $\mathcal M=(\mathbf X,\mathcal E)$ be an initial mesh with node coordinates $\mathbf X=[\mathbf x_1,\dots,\mathbf x_N]^\top\in\mathbb R^{N\times d}$ and connectivity $\mathcal E$. Let $u$ be one or multiple physical fields of interest on $\Omega$, and let $\mathbf U_X=[u(\mathbf x_1),\dots,u(\mathbf x_N)]^\top$ denote their nodal values sampled at the mesh nodes $\mathbf X$, and $\mathbf U_X$ is obtained by solving the PDE on the mesh $\mathcal M$ with a mesh-based numerical solver~\cite{ciarlet2002fem,brenner2008fem}. Learning-based mesh movement learns a mapping $\widehat{\mathbf X}=f_\theta(\mathbf X,\mathcal E,\mathbf U_X)$ that predicts new node coordinates while keeping the node count $N$ and the connectivity $\mathcal E$ unchanged. The resulting moved mesh is $\widehat{\mathcal M}=(\widehat{\mathbf X},\mathcal E)$. Its purpose is to yield a mesh on which $u$ can be solved more accurately, i.e., to improve the accuracy of the discrete physical field $\mathbf U_{\widehat X}$ associated with $\widehat{\mathcal M}$ relative to that on $\mathcal M$.

A monitor function provides the nodal density pattern that drives \(r\)-adaptation. For the mesh $\mathcal M$, it assigns a positive value to each node, $\mathbf m=[m_1,\dots,m_N]^\top\in\mathbb R_{>0}^N$, where a large $m_i$ indicates that node $i$ lies in a region where the mesh should be refined. Classical \(r\)-adaptation moves the nodes so that they are equidistributed with respect to $\mathbf m$, i.e., the mesh density follows the monitor values' distribution~\cite{mcrae2018optimal,budd2015geometry}. In this formulation, the monitor enters the movement mapping as an explicit conditioning input, $\widehat{\mathbf X}=f_\theta(\mathbf X,\mathcal E,\mathbf U_X,\mathbf m)$, so that $f_\theta$ learns to relocate nodes according to $\mathbf m$. The accuracy of the resulting field depends critically on whether $\mathbf m$ correctly identifies the important regions~\cite{fidkowski2021metric}.

In most prior learning-based \(r\)-adaptation methods, the monitor \(\mathbf m\) is typically constructed by heuristic rules or derived from local solution features, such as a Hessian-based monitor $m(\mathbf{x})=\sqrt{1+\alpha\|H(u)(\mathbf{x})\|_F^2}$~\cite{loseille2011continuous2}, a gradient-based monitor $m(\mathbf{x})=\sqrt{1+\alpha\|\nabla u(\mathbf{x})\|_2^2}$~\cite{zienkiewicz1992recovery}, or error estimators~\cite{ainsworth2000error}, and then held fixed during training. As illustrated in Figure~\ref{fig:teaser}, different monitor functions introduce different biases, e.g., gradient-based monitors favor steep regions with large first-order variation and can overemphasize linear slopes while underestimating smooth high-curvature regions. Hessian-based monitors favor regions with large second-order curvature and better capture curved layers and corners. Therefore, such bias leads to suboptimal accuracy and poor generalization across different PDEs.

\begin{figure*}[t]
    \centering
    \includegraphics[width=0.96\textwidth]{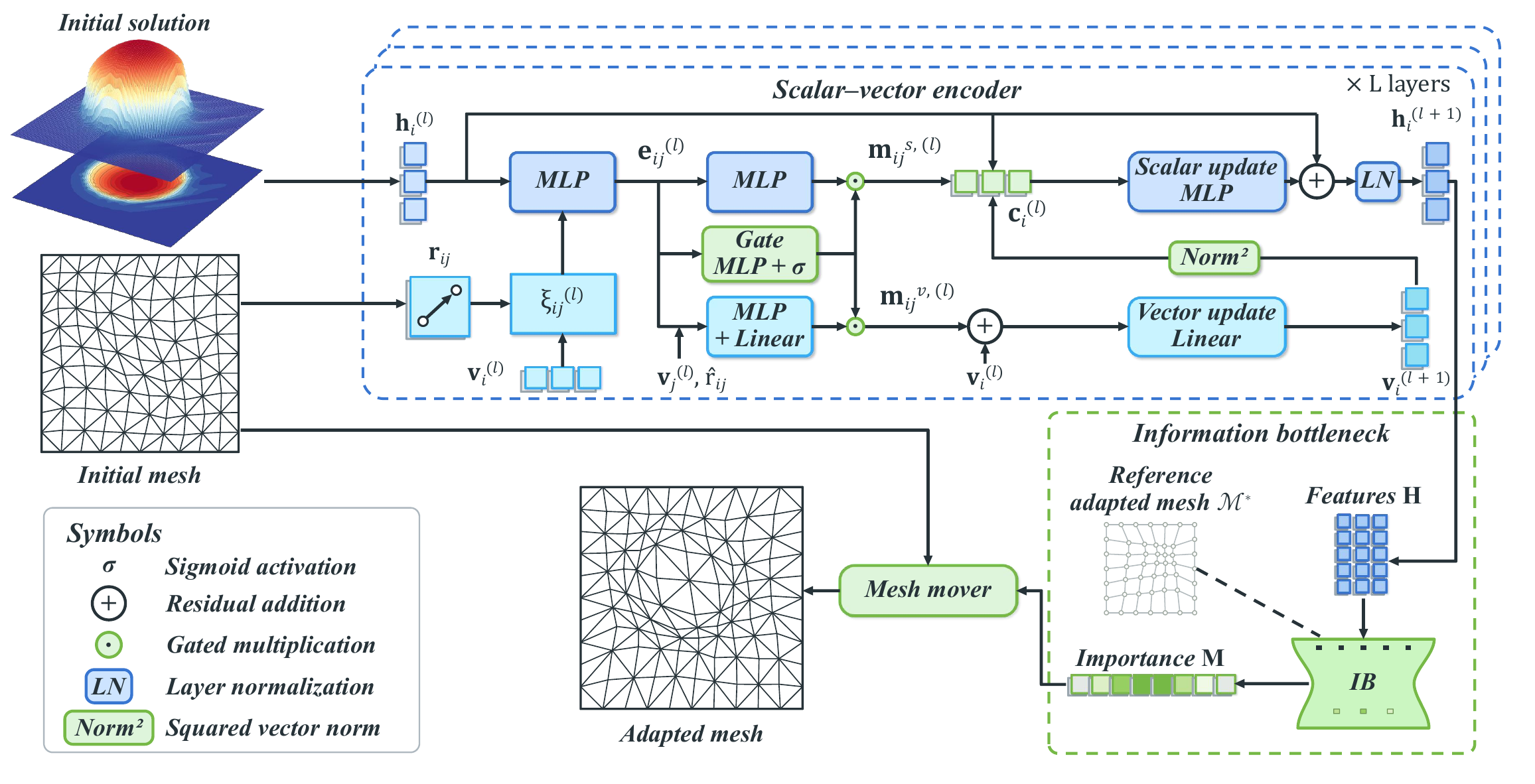}
    \caption{During training, PhyIB-Mesh jointly encodes the initial solution and mesh. A scalar–vector graph encoder maps them to discriminative node representations, from which an information-bottleneck module estimates node importance via mutual information with the reference mesh. These scores serve as a continuous monitor for a neural mesh mover to predict the adapted mesh, on which the numerical solution is computed by a mesh-based finite element method. After training, all parameters are frozen for mesh movement.}
    \label{fig:phyib_framework}
\end{figure*}

\section{Methodology}

As illustrated in Figure~\ref{fig:phyib_framework}, PhyIB-Mesh consists of two core modules. A scalar–vector state-based graph encoder first encodes the initial mesh and the observed physical field, yielding discriminative node representations that provide the information-bottleneck module with a basis for assessing node importance. An information-bottleneck module then quantifies node importance from these representations as node-wise scores, which serve as a continuous monitor for a neural mesh mover to predict the adapted mesh. 


\subsection{Graph Encoder with Scalar-Vector States}

The encoder fuses physical-field and mesh geometry into node representations: the field indicates regions of solution variation, while the geometry captures the current resolution distribution and stretching directions. Message passing updates a scalar state for field information and a vector-valued state for mesh information at each node and layer.

At the first layer, the scalar state is the nodal field value \(\mathbf{h}_i^{(0)}=\tilde u_i\mathbf{1}\), and the vector state is initialized as \(\mathbf{v}_i^{(0)}=\mathbf{0}\).
For each node \(i\), let \(\mathcal{N}(i)\) denote its neighborhood, constructed from the mesh connectivity. For an edge \(j\to i\), \(j\in\mathcal{N}(i)\), let \(\mathbf{r}_{ij}=\mathbf{x}_j-\mathbf{x}_i\), \(\widehat{\mathbf{r}}_{ij}=\frac{\mathbf{r}_{ij}}{\|\mathbf{r}_{ij}\|_2}\), where \(\widehat{\mathbf{r}}_{ij}\) is the unit edge direction. 
A geometric edge feature is first constructed to summarize how the two endpoint vector states align with the edge direction:
\begin{equation}
\boldsymbol{\xi}_{ij}^{(l)}
={}
\bar r_{ij}
\oplus
\bigl\langle \mathbf{v}_j^{(l)},\widehat{\mathbf{r}}_{ij}\bigr\rangle
\oplus
\bigl\langle \mathbf{v}_i^{(l)},\widehat{\mathbf{r}}_{ij}\bigr\rangle 
\oplus
\bigl\|\mathbf{v}_j^{(l)}\bigr\|_2
\oplus
\bigl\|\mathbf{v}_i^{(l)}\bigr\|_2,
\label{eq:edge_geom}
\end{equation}
where \(\oplus\) denotes concatenation, \(\langle\cdot,\cdot\rangle\) denotes the dot product, so that \(\langle\mathbf{v}_j^{(l)},\widehat{\mathbf{r}}_{ij}\rangle\) is the component of the vector state \(\mathbf{v}_j^{(l)}\) along the unit edge direction \(\widehat{\mathbf{r}}_{ij}\). This feature contains only geometric quantities, and it describes the local stretching and orientation of the mesh.

The above feature is then combined with the scalar states to form an edge encoding that fuses field information with geometry: $\mathbf{e}_{ij}^{(l)}
=
\mathrm{MLP}^{(l)}\left(
\mathbf{h}_j^{(l)}
\oplus
\mathbf{h}_i^{(l)}
\oplus
\boldsymbol{\xi}_{ij}^{(l)}
\right)$.
A sigmoid edge gate then produces a scalar weight for each edge: $a_{ij}^{(l)}
=
\sigma\left(\mathrm{MLP}^{(l)}(\mathbf{e}_{ij}^{(l)})\right)\in(0,1)$.
This weight scales the messages passed along the edge, so that edges that carry less useful physical or geometric information for the node update can be assigned smaller weights during training.

The edge encoding is then used to produce two messages from node \(j\) to node \(i\), both scaled by \(a_{ij}\): a scalar message $\mathbf{m}_{ij}^{s}$ that carries physical-field information and a vector message $\mathbf{m}_{ij}^{v}$ that carries directional geometric information:
\begin{equation}
\begin{aligned}
\mathbf{m}_{ij}^{s,(l)}
&=
a_{ij}^{(l)}
\mathrm{MLP}^{(l)}(\mathbf{e}_{ij}^{(l)}),
\\
\mathbf{m}_{ij}^{v,(l)}
&=
a_{ij}^{(l)}\Bigl[
\mathrm{MLP}^{(l)}(\mathbf{e}_{ij}^{(l)})\otimes\hat{\mathbf r}_{ij}+
\mathrm{Linear}^{(l)}(\mathbf{v}_j^{(l)})
\Bigr].
\end{aligned}
\label{eq:scalar_vector_msg}
\end{equation}
where \(\otimes\) denotes the multiplication of each predicted radial coefficient with the edge direction. For the vector message, the first term injects a learned radial component along the edge, and the second term transports the neighbor's vector state.
Using the two messages, the vector state is updated by aggregating incoming vector messages:
\begin{equation}
\begin{split}
\mathbf{v}_i^{(l+1)}
=
\mathrm{Linear}^{(l)}\Bigl(&
\mathbf{v}_i^{(l)}+
\frac{1}{|\mathcal{N}(i)|}
\sum_{j\in\mathcal{N}(i)}
\mathbf{m}_{ij}^{v,(l)}\Bigr).
\end{split}
\label{eq:vector_update}
\end{equation}
Using the updated vector state, one aggregated feature can be formed as,
\begin{equation}
\mathbf{c}_i^{(l)}
=
\mathbf{h}_i^{(l)}
\oplus
\frac{1}{|\mathcal{N}(i)|}
\sum_{j\in\mathcal{N}(i)}
\mathbf{m}_{ij}^{s,(l)}
\oplus
\left\|\mathbf{v}_i^{(l+1)}\right\|_2^2.
\label{eq:scalar_update_feature}
\end{equation}
This aggregates the field state and neighboring messages. Using aggregated features, the scalar state is  updated as
\begin{equation}
\mathbf{h}_i^{(l+1)}
=
\mathrm{LN}\left(
\mathbf{h}_i^{(l)}
+
\mathrm{MLP}^{(l)}(\mathbf{c}_i^{(l)})
\right),
\label{eq:scalar_update}
\end{equation}
where \(\mathrm{LN}\) denotes layer normalization~\cite{ba2016layernorm}. We stack \(L\) such message-passing layers, and the final node representation is \(\mathbf{h}_i^{(L)}\). The discriminative node representations encoding both physical-field and mesh geometric information provide the information-bottleneck module with a basis for computing node importance.

\subsection{Information Bottleneck-Based Monitor}
Given the node representations $\mathbf{H}=[\mathbf{h}_1,\dots,\mathbf{h}_N]^\top\in\mathbb{R}^{N\times d}$, our goal is to compute an importance value for each node with respect to a reference adapted mesh \(\mathcal{M}^*\). We introduce a continuous mask vector \(\mathbf{M}\in[0,1]^{N}\), where \(M_i\in[0,1]\) is the importance of node \(i\). 
A node with large \(M_i\) is treated as important for reproducing \(\mathcal{M}^*\).
Since \(r\)-adaptation keeps the node count fixed, refining one region must require coarsening another; if importance were high everywhere, no region would remain coarsenable. Therefore, we want $\mathbf{M}$ to be maximally informative about the mesh \(\mathcal{M}^*\), while keeping the importance distribution sparse so that only a small subset of nodes receives large importance values. This trade-off can be formulated as an information bottleneck~\cite{tishby2000information,alemi2017dvib},
\begin{equation}
\max\; - I(\mathbf{H},\mathbf{M})+\beta
I({\mathbf{M}},\mathcal{M}^*).
\end{equation}
The first term is the compression term, which penalizes the mask information about the full node set, and pushes mask values toward sparsity; the second term is the mesh fidelity term that preserves adapted mesh-relevant information.

\textbf{Compression term.}
We first derive a tractable form of the compression term, i.e., \(I(\mathbf{H},\mathbf{M})\). This mutual information involves the marginal \(p(\mathbf{M})\), which is intractable. We therefore introduce a sparse prior \(r(\mathbf{M})\) as a tractable surrogate for \(p(\mathbf{M})\). To parameterize the mask distribution, we use a mask generator
\(q_\theta(\mathbf{M}\mid\mathbf{H})=\prod_i\mathrm{Bernoulli}(M_i;p_i)\),
where \(p_i=\sigma(f_\theta(\mathbf{h}_i))\in(0,1)\) is the Bernoulli probability and is used directly as a continuous soft mask, and \(f_\theta\) is a two-layer MLP with parameters \(\theta\).
By the variational upper bound (see Appendix~\ref{app:ib_derivation} for the detailed derivation),
\begin{equation}
I(\mathbf{H},\mathbf{M})
\;\le\;
\mathbb{E}_{p(\mathbf{H})}
\mathcal{D}_{\mathrm{KL}}\!\left(q_\theta(\mathbf{M}\mid\mathbf{H})\,\|\,r(\mathbf{M})\right).
\end{equation}
Because both \(q_\theta\) and \(r\) are independent Bernoulli distributions, the KL has a closed form,
\begin{equation}\label{eq:klsp}
\begin{aligned}
&\mathcal{D}_{\mathrm{KL}}\!\left(q_\theta(\mathbf{M}\mid\mathbf{H})\,\|\,r(\mathbf{M})\right)
\\
&\quad=
\sum_{i=1}^{N}
\left[
p_i\log\frac{p_i}{\pi}
+
(1-p_i)\log\frac{1-p_i}{1-\pi}
\right].
\end{aligned}
\end{equation}
To encourage only a small subset of nodes to receive large mask values, we use a sparse prior with probability \(\pi=0.1\).
In training, the sparsity regularization \(\mathcal{L}_{\mathrm{KL}}\) is the dataset average of this single-sample KL,
\begin{equation}
\mathcal{L}_{\mathrm{KL}}
=
\mathbb{E}_{(\mathbf{H},\mathbf{M})\sim\mathcal{D}}
\left[
\mathcal{D}_{\mathrm{KL}}\!\left(q_\theta(\mathbf{M}\mid\mathbf{H})\,\|\,r(\mathbf{M})\right)
\right].
\end{equation}
This term penalizes mask values that deviate from the small prior \(\pi\): a node whose \(M_i\) stays close to \(\pi\) keeps a small importance value, whereas a node whose \(M_i\) moves toward one receives a large importance value at a larger penalty.

\textbf{Mesh fidelity term.}
To lower-bound \(I({\mathbf{M}},\mathcal{M}^*)\), we expand it as
\begin{equation}
I({\mathbf{M}},\mathcal{M}^*)
=
\mathbb{E}\big[\log p(\mathcal{M}^*\mid{\mathbf{M}})\big]
+
H(\mathcal{M}^*),
\end{equation}
where \(H(\mathcal{M}^*)\) is the entropy of the reference adapted mesh and is constant. The true posterior \(p(\mathcal{M}^*\mid{\mathbf{M}})\) is intractable, so we introduce a variational distribution \(q_\phi(\mathcal{M}^*\mid{\mathbf{M}})\).
Here \(q_\phi(\mathcal{M}^*\mid\mathbf{M})\) takes the form of predicting the adapted mesh \(\mathcal{M}^*\) from the monitor values \(\mathbf{M}\). This is exactly the form of any learning-based mesh mover that relies on monitor values.
By the nonnegativity of the KL divergence,
\begin{equation}
\mathbb{E}\big[\log p(\mathcal{M}^*\mid {\mathbf{M}})\big]
\ge
\mathbb{E}\big[\log q_\phi(\mathcal{M}^*\mid{\mathbf{M}})\big],
\end{equation}
which yields the variational lower bound
\begin{equation}
I({\mathbf{M}},\mathcal{M}^*)
\ge
\mathbb{E}\big[\log q_\phi(\mathcal{M}^*\mid{\mathbf{M}})\big]
+
H(\mathcal{M}^*).
\end{equation}
Since \(H(\mathcal{M}^*)\) is constant, maximizing the lower bound reduces to minimizing \(\mathbb{E}[-\log q_\phi]\). 

We assume that the variational distribution \(q_\phi\) models the adapted mesh as the predicted mesh plus isotropic Gaussian noise, i.e., $\mathcal{M}^*=\widehat{\mathcal{M}}+\boldsymbol{\epsilon}$, $\boldsymbol{\epsilon}\sim\mathcal{N}(\mathbf{0},\mathbf{I})$.
Under this Gaussian model, the negative log-likelihood of \(q_\phi\) is
\(-\log q_\phi(\mathcal{M}^*\mid\mathbf{M})
=
\frac{1}{2}\|\mathcal{M}^*-\widehat{\mathcal{M}}\|_2^2+\mathrm{const}\),
so maximizing the likelihood is equivalent to minimizing the squared coordinate error between \(\mathcal{M}^*\) and \(\widehat{\mathcal{M}}\).
Since directly minimizing coordinate error is unstable, we replace it with the element-level volume loss in previous work~\cite{zhang2024um2n}
\begin{equation}
\mathcal{L}_{\mathrm{mesh}}
=
\mathbb{E}_{\mathcal{D}}
\left[
\frac{1}{N_e}
\sum_{i=1}^{N_e}
\left|\mathrm{Vol}(\hat{t}_i)-\mathrm{Vol}(t_i^*)\right|
\right],
\end{equation}
where \(\hat{t}_i\) and \(t_i^*\) are the \(i\)-th elements of \(\widehat{\mathcal{M}}\) and \(\mathcal{M}^*\), respectively, and \(\mathrm{Vol}(\cdot)\) computes the element volume. It should be noted that \(\mathcal{L}_{\mathrm{mesh}}\) is not the negative log-likelihood \(
-\log q_\phi\), but this element-level volume loss that drives \(\widehat{\mathcal{M}}\) toward \(\mathcal{M}^*\) in the same way.

\textbf{Overall training objective.}
The training loss is the mesh fidelity term \(\mathcal{L}_{\mathrm{mesh}}\) and the compression term \(\mathcal{L}_{\mathrm{KL}}\). The optimization variables are the mask generator \(\theta\) and the mesh mover \(\phi\). The mask \(\mathbf{M}\) is computed from \(q_\theta\) in forward pass, i.e., \(M_i=\sigma(f_\theta(\mathbf{h}_i))\). Because \(q_\phi(\mathcal{M}^*\mid\mathbf{M})\) predicts the adapted mesh from the monitor values, it has the same form as a learning-based mesh mover that relies on monitor values. Hence, \(q_\phi\) can be instantiated by these movers. The overall objective is $\mathcal{L}_{\mathrm{total}}=\mathcal{L}_{\mathrm{mesh}}+\mathcal{L}_{\mathrm{KL}}$.

\textbf{Integrated with mesh mover.} Each mask entry \(M_i\) can be interpreted as the importance of node \(i\). Nodes that matter for the mesh movement retain large mask values. This is exactly the role played by a monitor function, which assigns each node a value describing the desired mesh density: large values indicate regions to be refined, small values indicate regions that can be coarsened. Therefore, these importance values serve directly as monitor function values and can be seamlessly integrated into mesh movement methods that rely on monitor function values.

PhyIB-Mesh follows the same solve--compute--move pipeline as the baseline mover~\cite{song2022m2n,zhang2024um2n}: starting from an initial mesh, the PDE is solved on the current mesh to obtain the physical field, a monitor is computed from this field, and the mesh is updated according to the monitor. Whether this pipeline is executed once or repeated in a loop, and whether the monitor is recomputed after each mesh update, follows the setting of the baseline mover being used. When the baseline movers repeat the pipeline, the monitor is recomputed from the updated physical field. PhyIB-Mesh does not change the baseline movers' schedule, and it only replaces the handcrafted monitor with a learned one.


\section{Experiments}

\subsection{Experimental Setup}

\begin{table*}[!t]
\caption{PDE ER (\%, $\uparrow$) for baselines and ablations.}
\label{tab:main_benchmark}
\centering
\small
\setlength{\tabcolsep}{2.0pt}
\renewcommand{\arraystretch}{1.10}
\resizebox{\textwidth}{!}{%
\begin{tabular}{lccccccc}
\toprule
Method / variant & Helmholtz & Swirl & Circular & SCF & L-shape & 3D Helmholtz & 3D OOD Helmholtz\\
\midrule
MA~\cite{budd2009moving} & $18.02$ & $55.33$ & $-10.45$ & \textit{Fail} (20/120) & $-4.06$ & $9.38$ & $13.09$\\
M2N~\cite{song2022m2n}  & $18.77\pm0.10$ & $15.43\pm11.74$ & \textit{Fail} (67/90) & \textit{Fail} (0/360) & $3.55\pm15.42$ & $11.79\pm0.34$ & $17.67\pm0.62$\\
UM2N~\cite{zhang2024um2n}  & $18.60\pm0.75$ & $53.82\pm6.70$ & $24.08\pm7.66$ & \textit{Fail} (346/360) & $3.34\pm3.73$ & $12.81\pm0.63$ & $19.42\pm1.05$\\
UGM2N~\cite{wang2025ugm2n}  & \textit{Fail} (296/303) & $24.89\pm2.70$ & \textit{Fail} (5/90) & \textit{Fail} (199/360) & \textit{Fail} (13/15) & $-2.27\pm0.71$ & $0.93\pm0.49$\\
G-adaptivity~\cite{rowbottom2025gadaptivity}  & $-14.98\pm6.85$ & $52.46\pm7.98$ & $14.46\pm18.53$ & $17.55\pm8.79$ & $-0.21\pm14.09$ & $1.52\pm0.50$ & $4.05\pm0.73$\\
DMM~\cite{hu2024dmm}  & \textit{Fail} (292/303) & \textit{Fail} (170/303) & $6.90\pm2.37$ & \textit{Fail} (193/360) & \textit{Fail} (0/15) & $0.03\pm0.03$ & $-0.04\pm0.06$\\
\midrule
\multicolumn{8}{l}{\emph{Objective ablations}}\\
w/o $\mathcal{L}_{\mathrm{kl}}$ & $31.18\pm0.07$ & $45.31\pm1.36$ & $34.97\pm2.65$ & $-17.65\pm18.32$ & $18.86\pm8.30$ & $34.81\pm0.15$ & $42.17\pm0.47$\\
w/o $\mathcal{L}_{\mathrm{mesh}}$ & $\mathbf{33.56\pm1.41}$ & $62.76\pm5.22$ & \textit{Fail} (50/90) & \textit{Fail} (237/360) & $22.07\pm7.12$ & $25.23\pm5.87$ & $31.31\pm10.57$\\
\midrule
\multicolumn{8}{l}{\emph{Graph encoder ablations}}\\
Graph Transformer~\cite{zhang2024um2n}  & $21.59\pm0.79$ & $29.02\pm32.64$ & $-12.54\pm21.93$ & $-13.21\pm11.35$ & $2.98\pm11.81$ & $31.83\pm0.72$ & $40.43\pm0.70$\\
Scalar-only & $18.66\pm1.63$ & $60.08\pm6.31$ & $25.48\pm6.58$ & $14.08\pm2.32$ & $1.57\pm3.64$ & $32.91\pm1.19$ & $40.32\pm1.33$\\
Mesh-only & $22.14\pm0.83$ & $61.39\pm4.95$ & $38.53\pm12.48$ & \textit{Fail} (332/360) & $-9.04\pm17.66$ & $32.24\pm0.45$ & $41.60\pm0.50$\\
w/o gate & $29.18\pm0.52$ & $59.22\pm11.87$ & $29.80\pm1.98$ & \textit{Fail} (331/360) & $21.06\pm14.43$ & $34.85\pm0.67$ & $42.63\pm1.11$\\
\midrule
\textbf{PhyIB-Mesh} & $29.24\pm0.36$ & $\mathbf{64.80\pm3.37}$ & $\mathbf{42.25\pm10.94}$ & $\mathbf{21.57\pm4.44}$ & $\mathbf{36.09\pm3.67}$ & $\mathbf{35.82\pm0.37}$ & $\mathbf{44.17\pm1.62}$\\
\bottomrule
\end{tabular}%
}
\par\medskip
\caption{PDE ER (\%, $\uparrow$) before and after PhyIB integration with baseline movers. $\Delta$ denotes PhyIB ER minus Original ER.}
\label{tab:cross_backbone_phyib}
\centering
\small
\setlength{\tabcolsep}{2.0pt}
\renewcommand{\arraystretch}{1.10}
\resizebox{\textwidth}{!}{%
\begin{tabular}{llccccccc}
\toprule
Backbone & Metric & Helmholtz & Swirl & Circular & SCF & L-shape & 3D Helmholtz & 3D OOD Helmholtz\\
\midrule
& Original ER & $18.60\pm0.75$ & $53.82\pm6.70$ & $24.08\pm7.66$ & \textit{Fail} (346/360) & $3.34\pm3.73$ & $12.81\pm0.63$ & $19.42\pm1.05$\\
UM2N~\cite{zhang2024um2n} & PhyIB ER & $29.24\pm0.36$ & $64.80\pm3.37$ & $42.25\pm10.94$ & $21.57\pm4.44$ & $36.09\pm3.67$ & $35.82\pm0.37$ & $44.17\pm1.62$\\
& $\Delta$ & $+10.64$ & $+10.98$ & $+18.17$ & \textit{Fail}$\to$\textit{valid} & $+32.75$ & $+23.01$ & $+24.75$\\
\midrule
& Original ER & $18.77\pm0.10$ & $15.43\pm11.74$ & \textit{Fail} (67/90) & \textit{Fail} (0/360) & $3.55\pm15.42$ & $11.79\pm0.34$ & $17.67\pm0.62$\\
M2N~\cite{song2022m2n} & PhyIB ER & $28.33\pm0.50$ & $42.21\pm1.01$ & $44.12\pm4.84$ & $39.34\pm5.78$ & $21.13\pm5.72$ & $35.87\pm0.34$ & $44.65\pm1.35$\\
& $\Delta$ & $+9.56$ & $+26.78$ & \textit{Fail}$\to$\textit{valid} & \textit{Fail}$\to$\textit{valid} & $+17.58$ & $+24.08$ & $+26.99$\\
\midrule
& Original ER & \textit{Fail} (296/303) & $24.89\pm2.70$ & \textit{Fail} (5/90) & \textit{Fail} (199/360) & \textit{Fail} (13/15) & $-2.27\pm0.71$ & $0.93\pm0.49$\\
UGM2N~\cite{wang2025ugm2n} & PhyIB ER & $18.67\pm0.04$ & $27.08\pm5.57$ & \textit{Fail} (70/90) & $23.93\pm9.41$ & $14.77\pm13.77$ & $19.23\pm0.10$ & $21.98\pm0.26$\\
& $\Delta$ & \textit{Fail}$\to$\textit{valid} & $+2.19$ & \textit{Fail}$\to$\textit{Fail} & \textit{Fail}$\to$\textit{valid} & \textit{Fail}$\to$\textit{valid} & $+21.50$ & $+21.05$\\
\midrule
& Original ER & $-14.98\pm6.85$ & $52.46\pm7.98$ & $14.46\pm18.53$ & $17.55\pm8.79$ & $-0.21\pm14.09$ & $1.52\pm0.50$ & $4.05\pm0.73$\\
G-adaptivity~\cite{rowbottom2025gadaptivity} & PhyIB ER & $22.19\pm0.10$ & $41.96\pm3.64$ & $40.60\pm3.49$ & $11.50\pm8.94$ & $24.15\pm6.27$ & $17.87\pm0.31$ & $25.13\pm1.04$\\
& $\Delta$ & $+37.17$ & $-10.50$ & $+26.15$ & $-6.05$ & $+24.36$ & $+16.36$ & $+21.07$\\
\bottomrule
\end{tabular}%
}
\end{table*}

\begin{figure*}[t]
\centering
\begin{minipage}[c]{.03\textwidth}
\centering\rotatebox{90}{\small\textbf{(a) Circular Layer}}
\end{minipage}\hfill
\begin{minipage}[c]{.90\textwidth}
\includegraphics[width=\linewidth]{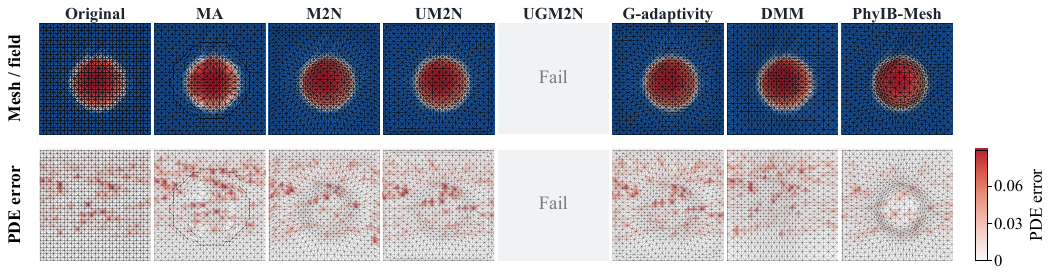}
\end{minipage}

\vspace{0.35em}
\begin{minipage}[c]{.03\textwidth}
\centering\rotatebox{90}{\small\textbf{(b) L-shape}}
\end{minipage}\hfill
\begin{minipage}[c]{.90\textwidth}
\includegraphics[width=\linewidth]{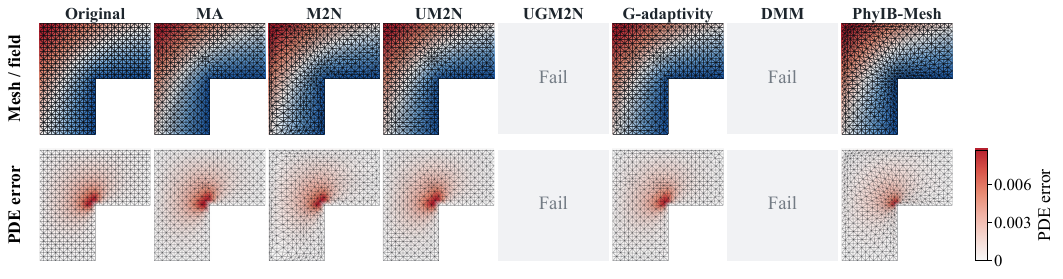}
\end{minipage}
\caption{The original/adapted mesh, the corresponding solution fields, and PDE errors on these meshes; “Fail” denotes mesh tangling.}
\label{fig:circular_lshape_mesh}
\end{figure*}

\begin{figure*}[t]
\centering
\includegraphics[width=0.90\textwidth]{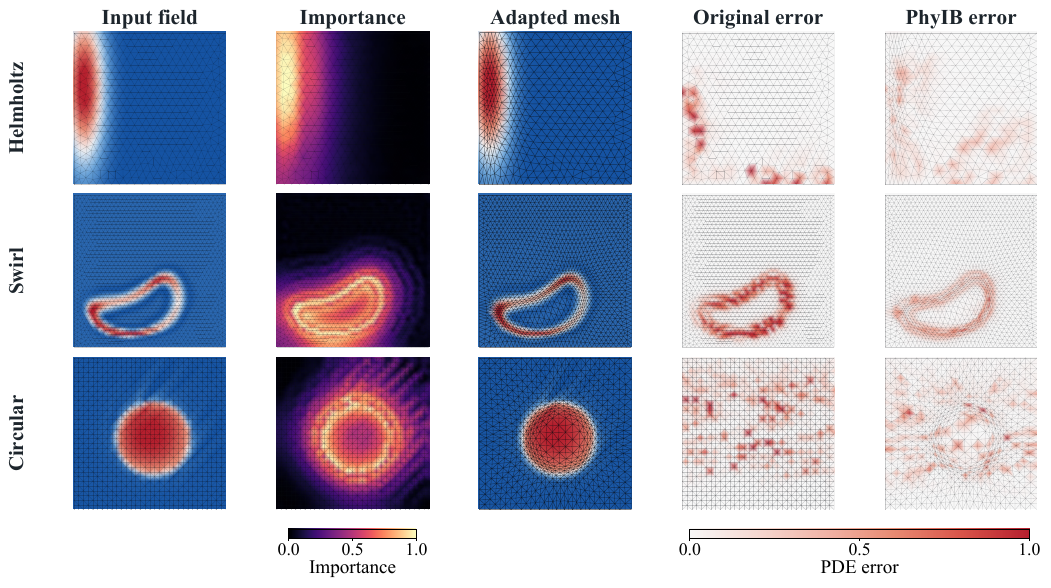}
\caption{Columns show the input mesh and its corresponding field, importance distribution, adapted mesh and its corresponding field, original-mesh error, and PhyIB-Mesh error. Importance and error colors are normalized within each row.}
\label{fig:ib_multibenchmark}
\end{figure*}

\paragraph{Benchmark.} We evaluate on seven systems: Helmholtz, Swirl, Circular Layer, Standing Contact Fatigue (SCF), and an L-shaped domain in 2D, together with 3D Helmholtz and 3D OOD Helmholtz. SCF involves predicting multiple physical fields, the L-shaped domain has an irregular boundary, and Swirl involves temporal evolution. 
Detailed benchmark descriptions are provided in Appendix~\ref{app:data}.

\paragraph{Baselines.} We compare the proposed method with traditional Monge-Ampère (MA) mesh movement~\cite{budd2009moving} and five learning-based methods: M2N~\cite{song2022m2n}, UM2N~\cite{zhang2024um2n}, UGM2N~\cite{wang2025ugm2n}, G-adaptivity~\cite{rowbottom2025gadaptivity}, and DMM~\cite{hu2024dmm}. Since our method can be seamlessly integrated into mesh movement methods that rely on monitor function values, we integrate it into UM2N, M2N, UGM2N, and G-adaptivity. All methods follow the same mesh budget. 
Following UM2N~\cite{zhang2024um2n}, our method does not require case-by-case retraining, thereby demonstrating its zero-shot generalization. 
We train separate 2D and 3D models using datasets containing a total of 600 and 300 randomly generated generic solution fields and their corresponding meshes, respectively. The adapted reference meshes are generated by MA from the initial meshes. At inference time, the proposed monitor functions generalize to other PDEs without requiring handcrafted monitors or MA-generated reference meshes.
Detailed training settings and implementation details are provided in Appendix~\ref{app:implementation}.

\paragraph{Metrics.} We use the PDE solution error reduction (ER). The PDE solution error is the \(L^2\)-norm between the solution on an obtained mesh and an accurate solution on a very high-resolution mesh. The ER measures how much the PDE solution error is reduced by mesh adaptation. Specifically, it is the difference between the PDE error on the original mesh and that on the adapted mesh.
To determine whether a mesh is valid, mesh tangling must be checked. Once mesh tangling occurs, the mesh becomes invalid for the numerical solver, which causes the simulation to break down. Parentheses indicate valid/total evaluations.

\subsection{Generalization performance}

Table~\ref{tab:main_benchmark} reports the ER of our full model PhyIB-Mesh, all baselines, and ablations across the seven benchmarks. Since our method is designed as a plug-in module for monitor-function-based mesh movement, we first integrate it into UM2N and report the resulting model as PhyIB-Mesh.
PhyIB-Mesh achieves the highest ER among all baselines on all seven benchmarks. On the 2D benchmarks, it improves ER substantially over the strongest baseline. On SCF, where baselines either suffer from mesh tangling or yield low ER, PhyIB-Mesh remains valid and attains an ER of 21.57\%. On the 3D benchmarks, it also outperforms UM2N and all other baselines by a large margin. 
Figure~\ref{fig:circular_lshape_mesh} further shows the original mesh, the meshes generated by different methods, and the corresponding solution fields and PDE solution errors obtained on these meshes. 

Despite using MA-generated references for supervision, our method can outperform MA because these references only provide generic element-volume targets. The network learns a field- and geometry-dependent importance mapping, optimized end-to-end with KL sparsity. At inference, it computes the monitor directly for downstream PDEs without MA-generated meshes. Because this monitor is input-dependent rather than fixed to MA’s handcrafted monitor, it can yield different density distributions and lower downstream errors.

Due to space limitations, more visualizations are provided in Appendix~\ref{app:qualitative}. Appendix~\ref{app:swirl_sequence} reports the temporal evolution of the dynamic fields and physical indicators on the adapted meshes, Appendix~\ref{app:cylinder_diagnostics} presents cylinder-wake visualizations and drag/lift coefficients, and Appendix~\ref{app:boundary_layer} examines whether adaptation resolves the rapid velocity change near the no-slip walls in parallel-plate flow. Parameter counts and latency are reported in Appendix~\ref{app:complexity}, where PhyIB-Mesh runs at approximately 14.3 ms with 2.91M parameters, comparable to existing learning-based methods.

\subsection{Ablation Studies}


In Table~\ref{tab:main_benchmark}, removing the KL term (w/o \(\mathcal{L}_{\mathrm{kl}}\)) degrades SCF and L-shape and reduces ER on most other benchmarks.
This shows that the KL term is critical for stable and accurate adaptation. Removing the element-volume loss (w/o $\mathcal{L}_{\mathrm{mesh}}$) causes mesh tangling on Circular Layer and SCF and lowers ER on L-shape and both 3D benchmarks. This confirms its role in maintaining valid meshes.

For the encoder, we specifically design a graph encoder that encodes both scalar and vector states. Compared with the Graph Transformer encoder used in UM2N~\cite{zhang2024um2n}, our designed encoder yields clear improvements. In addition, scalar-only (keeping the physical-field state information) and mesh-only (keeping the mesh geometry's vector state information) both degrade performance. In particular, Mesh-only fails on SCF and yields negative ER on L-shape, while Scalar-only achieves only 1.57\% on L-shape. This indicates that both physical-field and geometric inputs are necessary. Removing the learned edge gates (w/o gate) also causes SCF to fail and reduces L-shape ER to 21.06\%. Overall, among the ablation variants, PhyIB-Mesh achieves the best ER on six of the seven benchmarks while maintaining valid meshes. Sensitivity to the sparse prior in Eq. (\ref{eq:klsp}) is reported in Appendix~\ref{app:prior_sensitivity}. 


\begin{figure}[!t]
\centering
\includegraphics[width=0.98\linewidth]{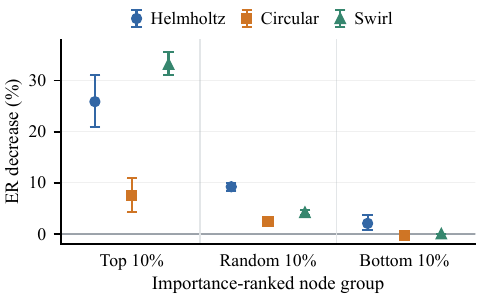}
\caption{ER decrease (\%) for perturbations of the obtained monitor values. Bars show 95\% confidence intervals.}
\label{fig:ib_evidence}
\vspace{-0.3cm}
\end{figure}

\subsection{Integration with Baseline Movers}

Table~\ref{tab:cross_backbone_phyib} reports ER before and after integrating our method into UM2N, M2N, UGM2N, and G-adaptivity. Integration improves ER on most benchmarks and converts many tangling-induced failures into valid evaluations. For UM2N, ER increases on all benchmarks, and SCF changes from failure to valid. For M2N, ER improves on Helmholtz, Swirl, L-shape, and both 3D benchmarks, while Circular Layer and SCF become valid. For UGM2N, Helmholtz, SCF, and L-shape change from failure to valid, and both 3D benchmarks improve substantially, although Circular Layer remains failed. For G-adaptivity, integration improves most benchmarks but slightly reduces ER on Swirl and SCF. Overall, our method is broadly compatible with different monitor-function-based backbones and typically delivers large gains.

\subsection{Investigation of Learned Monitor Values}
\label{sec:interpretability}

We evaluate whether the learned importance scores, i.e., monitor values, are physically meaningful. Figure~\ref{fig:ib_multibenchmark} shows the distribution of monitor values over the nodes. The values concentrate on regions with large physical gradients, as well as on the interior or surrounding areas enclosed by high-gradient regions. Large gradients indicate rapid local variation of the solution, where discretization errors are typically large, and mesh resolution should be increased. In addition, regions enclosed by or surrounding high-gradient zones often exhibit large curvature, and the solution there can be affected by error propagation from sharp layers. The adapted meshes generated from these monitor values yield lower simulation error than the original meshes. 


We further quantitatively analyze the learned monitor values. Specifically, we decrease the monitor values by \(0.01\) at the top \(10\%\), bottom \(10\%\), or a randomly selected \(10\%\) of the nodes. As shown in Figure~\ref{fig:ib_evidence}, perturbing the top-ranked group produces the largest average decrease in ER, whereas perturbing low-importance nodes has the smallest effect. This supports interpreting the learned scores as indicators of numerical sensitivity, since changing the monitor value at highly ranked nodes has a greater impact on the accuracy of the adapted solution.

\section{Conclusions}
We introduced PhyIB-Mesh, an information-bottleneck-based monitor function that learns node-wise importance values for mesh movement. PhyIB-Mesh learns the monitor from the physical field and mesh geometry through a geometry-physics aware encoder and an information-bottleneck module. The learned importance values can be seamlessly integrated into mesh movement methods that rely on monitor function values. Experiments across seven two- and three-dimensional benchmarks show that PhyIB-Mesh achieves competitive numerical accuracy and zero-shot generalization. The learned importance scores also align with nodes that have a strong influence on the PDE solution, supporting their use as meaningful monitor values.

\section*{LLM Usage Statement}
OpenAI Codex was used for language polishing and assisted code generation. The authors are responsible for the final manuscript, code verification, and the validity of the reported results.

{\small
\bibliographystyle{ieeenat_fullname_citation}
\bibliography{references}
}

\clearpage

\appendix
\titlespacing*{\subsection}{0pt}{3.25ex plus 1ex minus .2ex}{1.5ex plus .2ex}
\onecolumn
\setcounter{topnumber}{99}
\setcounter{totalnumber}{99}
\renewcommand{\topfraction}{0.99}
\renewcommand{\bottomfraction}{0.99}
\renewcommand{\textfraction}{0.01}
\begingroup
\centering
\Large
\textbf{\phyibpapertitle}\\
\vspace{0.5em}Supplementary Material \\
\vspace{1.0em}
\par
\endgroup

\section{Variational Inference on the Information Bottleneck}
\label{app:ib_derivation}

Recall that the information-bottleneck objective is
\begin{equation}
\max\; -I(\mathbf{H},\mathbf{M})
+\beta I(\mathbf{M},\mathcal{M}^*).
\end{equation}
where \(\mathbf{H}\) denotes the node representations, $\mathcal{M}^*$ is MA-generated reference adapted mesh, and
\(\mathbf{M}\) denotes the node-wise importance mask.
The compression term limits how much information the mask \(\mathbf{M}\) retains about the full node set \(\mathbf{H}\). By minimizing it, most mask values are pushed toward a sparse distribution, so that only a small subset of nodes receives large importance values. The mesh fidelity term rewards the mask for being informative about the reference adapted mesh \(\mathcal{M}^*\), i.e., the mask should preserve the information needed to reproduce the target mesh. Both terms are intractable in their exact forms: the first involves the marginal \(p(\mathbf{M})\), and the second involves the true posterior \(p(\mathcal{M}^*\mid\mathbf{M})\). We therefore bound both terms variationally and optimize the resulting tractable objective.

By definition,
\begin{equation}
I(\mathbf{H},\mathbf{M})
=
\mathbb{E}_{p(\mathbf{H},\mathbf{M})}
\left[
\log\frac{p(\mathbf{M}\mid\mathbf{H})}{p(\mathbf{M})}
\right].
\end{equation}
Under our model, the mask is produced by the mask generator, so
\(p(\mathbf{M}\mid\mathbf{H})=q_\theta(\mathbf{M}\mid\mathbf{H})=\prod_i\mathrm{Bernoulli}(M_i;p_i)\)
with \(p_i=\sigma(f_\theta(\mathbf{h}_i))\in(0,1)\), where \(p_i\) is the Bernoulli selection probability used directly as the continuous soft mask and \(f_\theta\) is a two-layer MLP with parameters \(\theta\). The marginal \(p(\mathbf{M})\) is intractable, so we introduce a sparse prior
\(r(\mathbf{M})=\prod_i\mathrm{Bernoulli}(M_i;\pi)\) and use it as a tractable surrogate for \(p(\mathbf{M})\). Splitting the logarithm gives
\begin{equation}
\begin{aligned}
I(\mathbf{H},\mathbf{M})
&=
\mathbb{E}_{p(\mathbf{H})q_\theta(\mathbf{M}\mid\mathbf{H})}
\left[
\log\frac{q_\theta(\mathbf{M}\mid\mathbf{H})}{r(\mathbf{M})}
\right] +
\mathbb{E}_{p(\mathbf{H})q_\theta(\mathbf{M}\mid\mathbf{H})}
\left[
\log\frac{r(\mathbf{M})}{p(\mathbf{M})}
\right] \\
&=
\mathbb{E}_{p(\mathbf{H})}
\mathcal{D}_{\mathrm{KL}}\!\left(q_\theta(\mathbf{M}\mid\mathbf{H})\,\|\,r(\mathbf{M})\right) -
\mathcal{D}_{\mathrm{KL}}\!\left(p(\mathbf{M})\,\|\,r(\mathbf{M})\right).
\end{aligned}
\end{equation}
The first equality follows by splitting the logarithm. The second equality recognizes the two expectations as KL divergences. Since the second KL term is nonnegative, we obtain the variational upper bound~\cite{alemi2017dvib,achille2018information}
\begin{equation}
I(\mathbf{H},\mathbf{M})
\;\le\;
\mathbb{E}_{p(\mathbf{H})}
\mathcal{D}_{\mathrm{KL}}\!\left(q_\theta(\mathbf{M}\mid\mathbf{H})\,\|\,r(\mathbf{M})\right).
\end{equation}
Because both \(q_\theta\) and \(r\) are independent Bernoulli distributions, the KL has a closed form,
\begin{equation}
\begin{aligned}
&\mathcal{D}_{\mathrm{KL}}\!\left(q_\theta(\mathbf{M}\mid\mathbf{H})\,\|\,r(\mathbf{M})\right)
\\
&\quad=
\sum_{i=1}^{N}
\left[
p_i\log\frac{p_i}{\pi}
+
(1-p_i)\log\frac{1-p_i}{1-\pi}
\right].
\end{aligned}
\end{equation}
To encourage only a small subset of nodes to receive large mask values, we set \(\pi=0.1\). In training, the sparsity regularization is the dataset average of this single-sample KL,
\begin{equation}
\mathcal{L}_{\mathrm{KL}}
=
\mathbb{E}_{(\mathbf{H},\mathbf{M})\sim\mathcal{D}}
\left[
\mathcal{D}_{\mathrm{KL}}\!\left(q_\theta(\mathbf{M}\mid\mathbf{H})\,\|\,r(\mathbf{M})\right)
\right].
\end{equation}
This term penalizes selection probabilities that deviate from the small prior \(\pi\): a node whose \(p_i\) stays close to \(\pi\) keeps a small importance value, whereas a node whose \(p_i\) moves toward one receives a large importance value at a larger penalty. In this way, the upper bound turns the intractable compression term into a tractable sparsity regularizer on the mask.

The derivation of the lower bound for the mesh fidelity term is provided in full in the main text. Starting from
\begin{equation}
I(\mathbf{M},\mathcal{M}^*)
=
\mathbb{E}\big[\log p(\mathcal{M}^*\mid\mathbf{M})\big]
+
H(\mathcal{M}^*),
\end{equation}
maximizing the lower bound reduces to minimizing \(\mathbb{E}[-\log q_\phi]\).
We assume that the variational distribution \(q_\phi\) models the adapted mesh as the predicted mesh plus isotropic Gaussian noise, i.e.,
\begin{equation}
\mathcal{M}^*=\widehat{\mathcal{M}}+\boldsymbol{\epsilon},
\qquad
\boldsymbol{\epsilon}\sim\mathcal{N}(\mathbf{0},\mathbf{I}).
\end{equation}
Under this Gaussian model, the negative log-likelihood of \(q_\phi\) is
\begin{equation}
-\log q_\phi(\mathcal{M}^*\mid\mathbf{M})
=
\frac{1}{2}\|\mathcal{M}^*-\widehat{\mathcal{M}}\|_2^2+\mathrm{const},
\end{equation}
which pushes the predicted mesh toward the reference mesh. However, directly minimizing the coordinate error is unstable in practice, so we use an element-level volume loss as the surrogate loss:
\begin{equation}
\mathcal{L}_{\mathrm{mesh}}
=
\mathbb{E}_{\mathcal{D}}
\left[
\frac{1}{|\widehat{\mathcal{M}}|}
\sum_{i=1}^{|\widehat{\mathcal{M}}|}
\left|
\mathrm{Vol}(\hat{t}_i)-\mathrm{Vol}(t_i^*)
\right|
\right],
\end{equation}
where \(\hat{t}_i\) and \(t_i^*\) are the \(i\)-th elements of \(\widehat{\mathcal{M}}\) and \(\mathcal{M}^*\), respectively.

In summary, maximizing the information-bottleneck objective is equivalent to minimizing the overall loss
\begin{equation}
\mathcal{L}_{\mathrm{total}}
=
\mathcal{L}_{\mathrm{mesh}}
+
\mathcal{L}_{\mathrm{KL}}.
\end{equation}

\clearpage
\section{Datasets Details}
\label{app:data}
\nocite{john1997nonconforming,comsolStandingContactFatigue,comsolMultiphysics64,grisvard2011elliptic,klambauer2017selu}



\paragraph{Numerical software and reference meshes.}
The PDEs are solved with Firedrake 0.13.0~\cite{rathgeber2016firedrake} to obtain numerical solutions on the generated meshes.
Reference adapted meshes $\mathcal{M}^*$ are generated offline using Monge--Amp\`ere-based mesh movement. For the 2D training data, we follow the UM2N generation pipeline~\cite{zhang2024um2n}, using the relaxation method implemented in Movement; for the 3D data, we use a corresponding volumetric implementation. Signed triangle areas in 2D and tetrahedral volumes in 3D, computed from these reference meshes, provide supervision for the mesh-fidelity loss.


%

\paragraph{Training data.}
The datasets contain $600$ 2D and $300$ 3D samples in total, with splits listed in Table~\ref{tab:dataset_summary}. Following UM2N~\cite{zhang2024um2n}, we construct Gaussian fields on the unit square:
\begin{equation}
u_{\mathrm{exact}}(x,y)=\sum_{k=1}^{K}a_k
\exp\!\left[-\left(\frac{(x-\mu_{x,k})^2}{\sigma_{x,k}^2}
+\frac{(y-\mu_{y,k})^2}{\sigma_{y,k}^2}\right)\right],
\qquad 1\leq K\leq6.
\label{eq:app_generic_field}
\end{equation}
Here $u_{\mathrm{exact}}$ is the prescribed analytical field and $x,y$ are spatial coordinates. $K$ is the number of Gaussian components; $a_k$, $(\mu_{x,k},\mu_{y,k})$, and $(\sigma_{x,k},\sigma_{y,k})$ are the amplitude, center, and widths of component $k$. For 3D, we use
\begin{equation}
u_{\mathrm{exact}}(x,y,z)=\sum_{k=1}^{K}a_k
\exp\!\left[-\left(\frac{(x-\mu_{x,k})^2}{\sigma_{x,k}^2}
+\frac{(y-\mu_{y,k})^2}{\sigma_{y,k}^2}
+\frac{(z-\mu_{z,k})^2}{\sigma_{z,k}^2}\right)\right],
\label{eq:app_generic_field_3d}
\end{equation}
where $z$ is the third spatial coordinate and $\mu_{z,k},\sigma_{z,k}$ are the corresponding center and width. These fields define the forcing and Dirichlet data below.
For both 2D and 3D fields, $K$ is sampled uniformly from $\{1,\ldots,6\}$. Amplitudes and center coordinates are independently sampled from $\mathcal{U}(0,1)$. Coordinate widths are sampled from $\mathcal{N}(1/4,(1/6)^2)$, rounded to three decimal places, and lower-clipped at $1/8$. Amplitudes and centers are also rounded to three decimal places.

\begin{table}[htbp]
\caption{Data partitions and numerical evaluation sets; $h$ denotes the characteristic initial-mesh size. A dash denotes a downstream benchmark not used for training.}
\label{tab:dataset_summary}
\centering
\small
\begin{tabular}{lcccc}
\toprule
Dataset & Dimension & Train & Validation & Formal evaluation \\
\midrule
Generic fields, $h=0.05$ & 2D & 225 & 30 & 45 internal holdout \\
Generic fields, $h=0.055$ & 2D & 225 & 30 & 45 internal holdout \\
Helmholtz & 2D & -- & -- & 101 cases \\
Swirl OOD & 2D & -- & -- & 101 states \\
Circular Layer OOD & 2D & -- & -- & 30 cases \\
SCF multi-field OOD & 2D & -- & -- & 20 cases $\times$ 7 fields \\
L-shaped domain OOD & 2D & -- & -- & 5 resolutions \\
3D Helmholtz & 3D & 225 & 45 & 30 cases \\
3D OOD Helmholtz & 3D & -- & -- & 7 cases \\
\bottomrule
\end{tabular}
\end{table}

\paragraph{Helmholtz.}
We use $101$ manufactured Gaussian cases from UM2N~\cite{zhang2024um2n} on $\Omega=(0,1)^2$:
\begin{equation}
-\Delta u+u=f\quad\text{in }\Omega,
\qquad u=u_{\mathrm{exact}}\quad\text{on }\partial\Omega,
\label{eq:app_helmholtz_pde}
\end{equation}
Here $u$ is the scalar PDE solution, $f=-\Delta u_{\mathrm{exact}}+u_{\mathrm{exact}}$ the source term, and $\partial\Omega$ the boundary of the domain $\Omega$. We use continuous Galerkin finite elements of degree one (CG1) and compare with the analytical field on the common fine mesh.

\paragraph{Swirl.}
The Swirl benchmark from UM2N~\cite{zhang2024um2n} records $101$ states of scalar transport on $(0,1)^2$:
\begin{equation}
\partial_t u+\nabla\cdot(\mathbf b u)=0.
\label{eq:app_swirl_pde}
\end{equation}
Here $u$ is concentration, $\mathbf b$ the prescribed velocity, and $t$ time. The initial field $u_0$ and velocity are
\begin{equation}
u_0(x,y)=\exp\!\left[-\frac{\left(\sqrt{(x-x_0)^2+(y-y_0)^2}-r_0\right)^2}{2\sigma^2}\right],
\label{eq:app_swirl_initial}
\end{equation}
\begin{equation}
\mathbf b(x,y,t)=0.9\,a(t)
\big(\sin^2(\pi x)\sin(2\pi y),-\sin^2(\pi y)\sin(2\pi x)\big).
\label{eq:app_swirl_velocity}
\end{equation}
The ring has center $(x_0,y_0)=(0.3,0.3)$, radius $r_0=0.2$, and width $\sigma=0.05/3$; the reversal factor $a(t)=1$ before $t=0.5$ and $-1$ thereafter. Data generation uses discontinuous piecewise-linear (DG1) elements, and evaluation uses CG1, both with a third-order strong-stability-preserving Runge--Kutta scheme and time step $\Delta t=0.001$ over $t\in[0,1]$.

\paragraph{Circular Layer.}
We construct $30$ parameterized cases from the circular-layer problem~\cite{john1997nonconforming} on $(0,1)^2$, with zero Dirichlet data:
\begin{equation}
-\epsilon\Delta u+\mathbf b\cdot\nabla u+cu=f,
\qquad \mathbf b=(2,3),\quad c=2.
\label{eq:app_circular_pde}
\end{equation}
Here $\epsilon$ is diffusivity, $\mathbf b$ advection velocity, and $c$ the reaction coefficient. The manufactured solution is
\begin{equation}
u_{\mathrm{exact}}(x,y)=16x(1-x)y(1-y)\big[\tfrac12+\pi^{-1}\arctan(q)\big],
\label{eq:app_circular_exact}
\end{equation}
where $q=2\epsilon^{-1/2}[r^2-(x-c_x)^2-(y-c_y)^2]$ with layer center $(c_x,c_y)$ and radius $r$; $f$ follows by substitution. The cases vary $\epsilon\in[5\cdot10^{-5},3\cdot10^{-4}]$, $c_x,c_y\in[0.42,0.58]$, and $r\in[0.20,0.28]$. We use CG1 elements with streamline upwind/Petrov--Galerkin (SUPG) stabilization with $n=32$ subdivisions per coordinate.

\paragraph{SCF.}
Inspired by the COMSOL Standing Contact Fatigue example~\cite{comsolStandingContactFatigue}, we construct a simplified axisymmetric linear-elastic problem in Firedrake in radial--axial coordinates $(r,z)\in[0,5]^2$:
\begin{equation}
-\nabla\cdot\boldsymbol\sigma=\mathbf0,
\qquad \boldsymbol\sigma=\lambda\,\mathrm{tr}(\boldsymbol\varepsilon_{\mathrm{axi}})\mathbf I
+2\mu\boldsymbol\varepsilon_{\mathrm{axi}},
\label{eq:app_scf_pde}
\end{equation}
Here $\boldsymbol\sigma$ is stress, $\boldsymbol\varepsilon_{\mathrm{axi}}$ axisymmetric strain, $\lambda,\mu$ the Lam\'e parameters, and $\mathbf I$ the identity tensor. With Young's modulus $E=1$ and Poisson's ratio $0.3$, the top surface receives Gaussian pressure $p(r)$
\begin{equation}
p(r)=A\exp[-r^2/(2s^2)].
\label{eq:app_scf_load}
\end{equation}
The bottom is fixed, the axis has zero radial displacement $u_r=0$, and the outer radial surface is traction-free. The $20$ cases vary pressure amplitude $A\in[0.010,0.030]$ and width $s\in[0.35,0.70]$. We solve displacement with vector-valued quadratic (CG2) elements and project the output fields to CG1. The seven outputs comprise displacement, stress, and stress-derived plastic-strain/fatigue indicators. The plastic-strain indicator is zero throughout these cases, giving zero error before adaptation. ER and validity are reported for the other six fields ($360$ evaluations across three training seeds; $120$ for deterministic MA).

\paragraph{L-shape.}
On the L-shaped domain $\Omega_L=(-1,1)^2\setminus([0,1]\times[-1,0])$, we solve the corner-singularity problem~\cite{grisvard2011elliptic}:
\begin{equation}
-\Delta u=0\quad\text{in }\Omega_L,
\qquad u=r^{2/3}\sin(2\theta/3)\quad\text{on }\partial\Omega_L,
\label{eq:app_lshape_pde}
\end{equation}
Here $r,\theta$ are polar coordinates about the reentrant corner, with $0\leq\theta\leq3\pi/2$. We use CG1 elements at five resolutions $n\in\{20,24,28,32,36\}$ and compare with the analytical solution at shared spatial probes.

\paragraph{3D Helmholtz.}
We construct the 3D Helmholtz dataset in Firedrake on the unit cube $\Omega_3=(0,1)^3$. Each analytical field $u_{\mathrm{exact}}$ is a weighted sum of one to six randomly generated three-dimensional Gaussian components, as defined in Equation~\ref{eq:app_generic_field_3d}. The component amplitudes, centers, and widths along each coordinate direction are randomly sampled. These analytical fields determine the source term $f=-\Delta u_{\mathrm{exact}}+u_{\mathrm{exact}}$ and Dirichlet boundary data for the following PDE:
\begin{equation}
-\Delta u+u=f\quad\text{in }\Omega_3,
\qquad u=u_{\mathrm{exact}}\quad\text{on }\partial\Omega_3.
\label{eq:app_helmholtz3d_pde}
\end{equation}
Both 3D evaluations use initial meshes with $729$ nodes and $3{,}072$ tetrahedra. The $30$ held-out cases use CG1 tetrahedral elements, with errors measured against the analytical manufactured solution.

\paragraph{3D OOD Helmholtz.}
This OOD evaluation instead constructs each analytical field from a single Gaussian component, forming one localized peak whose position and spatial extent vary across cases. It uses the same PDE form as 3D Helmholtz in Equation~\ref{eq:app_helmholtz3d_pde}, with the source term and boundary values recomputed from each analytical field.

\clearpage
\section{Implementation Details}
\label{app:implementation}

The encoder receives the initial node coordinates, physical-field values, and mesh connectivity. The input field $\widetilde u_i$ is centered and normalized by its root-mean-square deviation within each sample.
The encoder uses six scalar--vector message-passing blocks with $96$ scalar channels and $16$ vector channels. Each edge encoder is a $257\!\rightarrow\!96\!\rightarrow\!96$ MLP with SELU activations~\cite{klambauer2017selu}. The sigmoid gate, scalar-message, and radial-vector heads use $96\!\rightarrow\!32\!\rightarrow\!1$, $96\!\rightarrow\!96\!\rightarrow\!96$, and $96\!\rightarrow\!32\!\rightarrow\!16$, respectively. The scalar-update MLP uses $208\!\rightarrow\!96\!\rightarrow\!96$ channels with a SELU hidden activation. The normalized edge length is $\bar r_{ij}=r_{ij}/\langle r\rangle_{\mathcal E}$, where $\langle r\rangle_{\mathcal E}$ is the mean over the encoder edges.

\begin{table}[htbp]
\caption{Configuration of the method components and default training settings.}
\label{tab:appendix_impl}
\centering
\small
\setlength{\tabcolsep}{4pt}
\begin{tabular}{ll}
\toprule
Setting & Configuration \\
\midrule
Scalar / vector channels & $96$ / $16$ \\
Message-passing blocks & $6$ \\
Edge encoder & $257\!\rightarrow\!96\!\rightarrow\!96$ \\
Edge-gate head & $96\!\rightarrow\!32\!\rightarrow\!1$ \\
Scalar-message head & $96\!\rightarrow\!96\!\rightarrow\!96$ \\
Radial-vector head & $96\!\rightarrow\!32\!\rightarrow\!16$ \\
Scalar-update MLP & $208\!\rightarrow\!96\!\rightarrow\!96$ \\
Importance-head hidden width & $32$ \\
Bernoulli prior $\pi$ & $0.1$ \\
Mesh-fidelity weight & $1.0$ \\
Optimizer & AdamW \\
Learning rate / weight decay & $10^{-4}$ / $10^{-6}$ \\
Training duration (maximum) & $1{,}000$ epochs \\
2D / 3D gradient accumulation & $7$ / $4$ samples \\
2D / 3D gradient clipping norms & $0.5$ / $1.0$ \\
Independent training seeds & $3$ \\
Training and inference monitor & $p_i=M_i$ \\
\bottomrule
\end{tabular}
\end{table}

We train separate 2D and 3D models on randomly generated generic solution fields and their corresponding meshes. Training uses AdamW with a learning rate of $10^{-4}$ on an NVIDIA RTX 3090 GPU. We use three independent training seeds for fair comparison. 



\clearpage
\section{Additional Experimental Results}
\label{app:additional_evidence}
\setlength{\intextsep}{6pt}
\setlength{\textfloatsep}{8pt}
\setlength{\floatsep}{6pt}

The additional experimental results comprise six parts. Section~\ref{app:qualitative} compares original and adapted meshes, solution fields, and PDE errors for Helmholtz, Swirl, and the 3D benchmarks. Section~\ref{app:swirl_sequence} shows how meshes and errors evolve over time. Section~\ref{app:cylinder_diagnostics} examines whether mesh adaptation resolves the cylinder wake and reproduces drag and lift responses. Section~\ref{app:boundary_layer} compares near-wall velocity profiles. Section~\ref{app:complexity} reports parameter counts and inference latency. Section~\ref{app:prior_sensitivity} tests sensitivity to the sparse prior $\pi$.




\subsection{More Mesh Visualizations}
\label{app:qualitative}

\begin{figure}[H]
\centering
\includegraphics[width=0.72\textwidth]{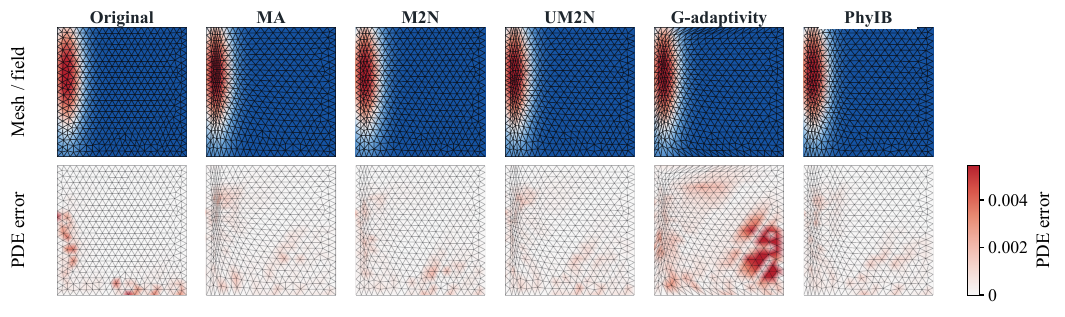}

\vspace{0.2em}
\includegraphics[width=0.72\textwidth]{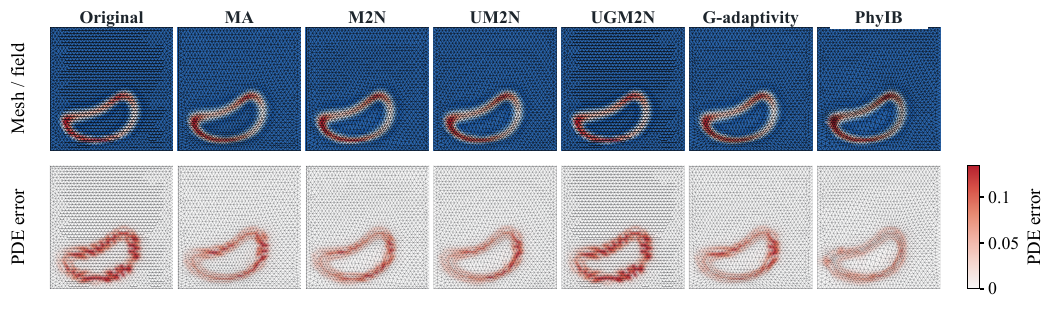}
\caption{\textbf{2D mesh and error comparisons.} Helmholtz (top block) and Swirl (bottom block); each block shows mesh/field views and corresponding pointwise absolute PDE errors.}
\label{fig:app_helmholtz_swirl_mesh}
\end{figure}

Figure~\ref{fig:app_helmholtz_swirl_mesh} compares meshes, solution fields, and pointwise PDE errors for representative Helmholtz and Swirl cases. PhyIB concentrates nodes around the localized Helmholtz peak and the thin Swirl structure. In Helmholtz, PhyIB reduces localized PDE errors and maintains a more uniform error distribution. In Swirl, the error band along the curved structure is weaker than for M2N, UM2N, and UGM2N. These comparisons connect the targeted node redistribution to improved local solution accuracy.

\begin{figure}[H]
\centering
\includegraphics[width=0.72\textwidth]{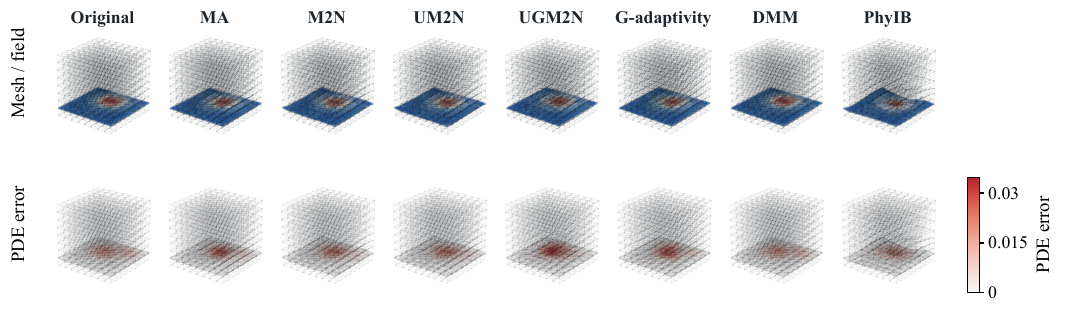}

\vspace{0.2em}
\includegraphics[width=0.72\textwidth]{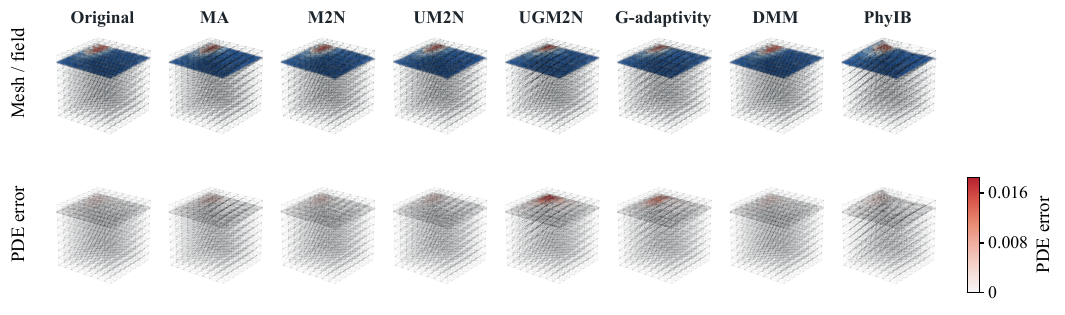}
\caption{\textbf{3D mesh and error comparisons.} 3D Helmholtz (top block) and 3D OOD Helmholtz (bottom block), shown on representative native slices with the surrounding tetrahedral mesh for context.}
\label{fig:app_3d_mesh_comparison}
\end{figure}

\Needspace{0.50\textheight}
Figure~\ref{fig:app_3d_mesh_comparison} shows the corresponding 3D Helmholtz and 3D OOD Helmholtz comparisons. PhyIB redistributes interior nodes toward the localized field structure. Its error slices show weaker hotspots than Original and the other baselines, particularly around the localized peak in the OOD case. Thus, the improved spatial resolution also reduces local solution errors on tetrahedral meshes.

\Needspace{0.32\textheight}
\subsection{Time-dependent PDE}
\label{app:swirl_sequence}
Swirl is time dependent, so a single snapshot cannot show whether mesh adaptation follows the evolving field throughout its deformation. We therefore visualize multiple times to assess whether PhyIB consistently tracks the transported structure and maintains solution accuracy as it stretches and recovers.

We compare Original, MA, M2N, UM2N, UGM2N, G-adaptivity, DMM and PhyIB at ten Swirl101 states with case indices $10,20,\ldots,100$, spanning $t=0.099$--$0.999$ s. The learned methods use seed 666. All methods share the fine-grid Firedrake reference~\cite{rathgeber2016firedrake}, coarse-grid solver and CG1 error projection. We check raw outputs for inversion and normal-boundary violations, allowing tangential sliding on the square boundary.

\begin{figure}[H]
\centering
\includegraphics[width=.75\linewidth,height=.87\textheight,keepaspectratio]{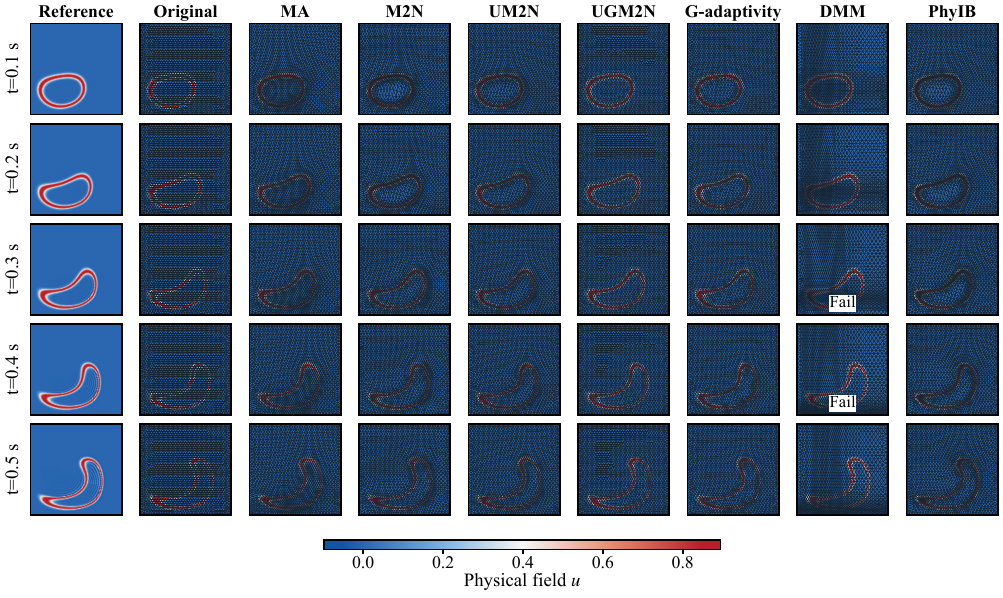}
\caption{Swirl mesh snapshots at $t\approx0.1$--$0.5$ s. Color denotes the reference physical field; lines show raw output meshes.}
\label{fig:app_swirl_mesh_early}
\end{figure}

\begin{figure}[H]
\centering
\includegraphics[width=.75\linewidth,height=.87\textheight,keepaspectratio]{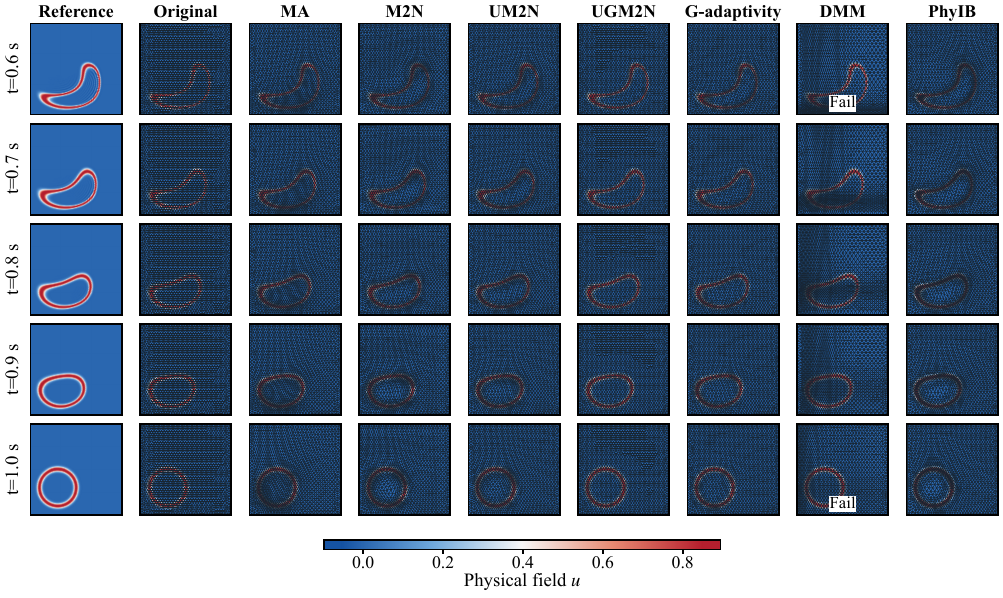}
\caption{Swirl mesh snapshots at $t\approx0.6$--$1.0$ s. Color denotes the reference physical field; lines show raw output meshes.}
\label{fig:app_swirl_mesh_late}
\end{figure}

\begin{figure}[H]
\centering
\includegraphics[width=.98\textwidth,height=.84\textheight,keepaspectratio]{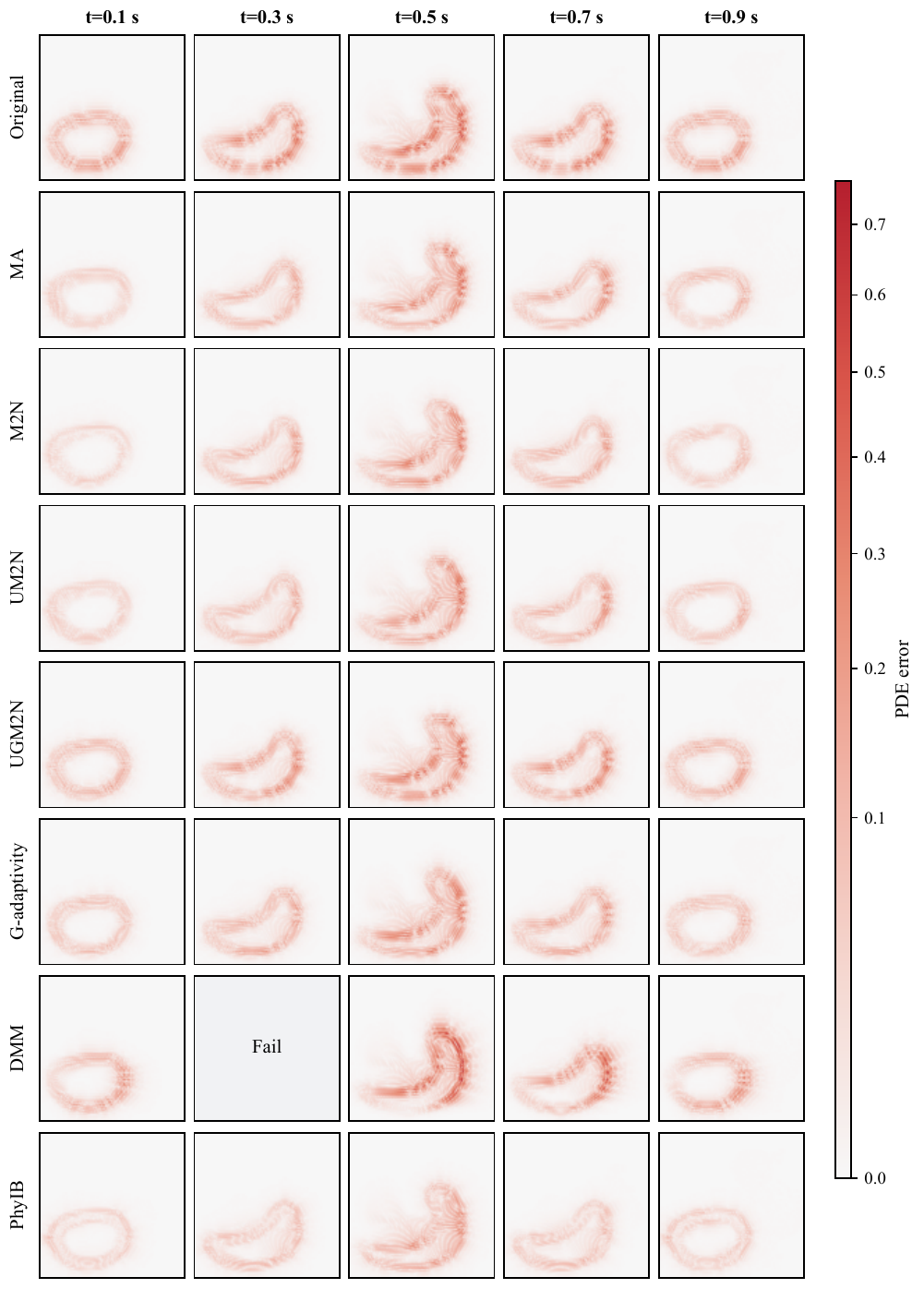}
\caption{Swirl absolute PDE-error fields at selected times, with a shared square-root color scale. Gray panels denote unavailable fields from invalid raw meshes.}
\label{fig:app_swirl_error_sequence}
\end{figure}

As shown in Figures~\ref{fig:app_swirl_mesh_early} and~\ref{fig:app_swirl_mesh_late}, the field stretches from a ring into a thin curved structure and then returns toward a ring. Unlike the fixed Original mesh, PhyIB moves its dense regions with the field, maintaining resolution along the thin curved structure during deformation. The colors show the reference field and indicate where resolution is needed. PhyIB also yields valid meshes at all ten selected states, whereas DMM is valid at six. Together with the error comparison in Figure~\ref{fig:app_swirl_error_sequence}, this demonstrates both field tracking and accurate solutions over the sampled evolution.


Figure~\ref{fig:app_swirl_error_sequence} compares absolute PDE errors at five times on a shared color scale. At $t\approx0.5$ s, when the field is most stretched, PhyIB shows a weaker high-error band along the thin structure than the other baselines. Its error concentrations are also reduced relative to Original at the earlier and later times. The benefit therefore persists through the deformation and recovery, rather than appearing only in a single snapshot. Gray panels denote invalid DMM outputs.

\clearpage
\subsection{Cylinder Flow and Force Visualization}
\label{app:cylinder_diagnostics}
We assess cylinder-wake resolution using vorticity fields and profiles, and evaluate the resulting drag and lift against a high-resolution reference.

Following UM2N~\cite{zhang2024um2n}, the flow satisfies
\begin{equation}
\partial_t\mathbf v+(\mathbf v\cdot\nabla)\mathbf v-\nu\Delta\mathbf v+\nabla p=\mathbf0,
\qquad \nabla\cdot\mathbf v=0.
\label{eq:app_cylinder_pde}
\end{equation}
Here $\mathbf v$ is velocity, $p$ pressure, and $\nu$ kinematic viscosity. The channel $(0,2.2)\times(0,0.41)$ contains a radius-$0.05$ cylinder centered at $(0.2,0.2)$. The inlet velocity is $\mathbf v_{\mathrm{in}}=(4(1.5)y(H_c-y)/H_c^2,0)$ with channel height $H_c=0.41$; the walls and cylinder are no-slip, the outlet has $p=0$, and $\nu=0.001$ (Reynolds number $\mathrm{Re}=100$). We use CG2 velocity and CG1 pressure elements with incremental pressure correction.

\begin{figure}[H]
\centering
\includegraphics[width=.84\textwidth,height=.86\textheight,keepaspectratio]{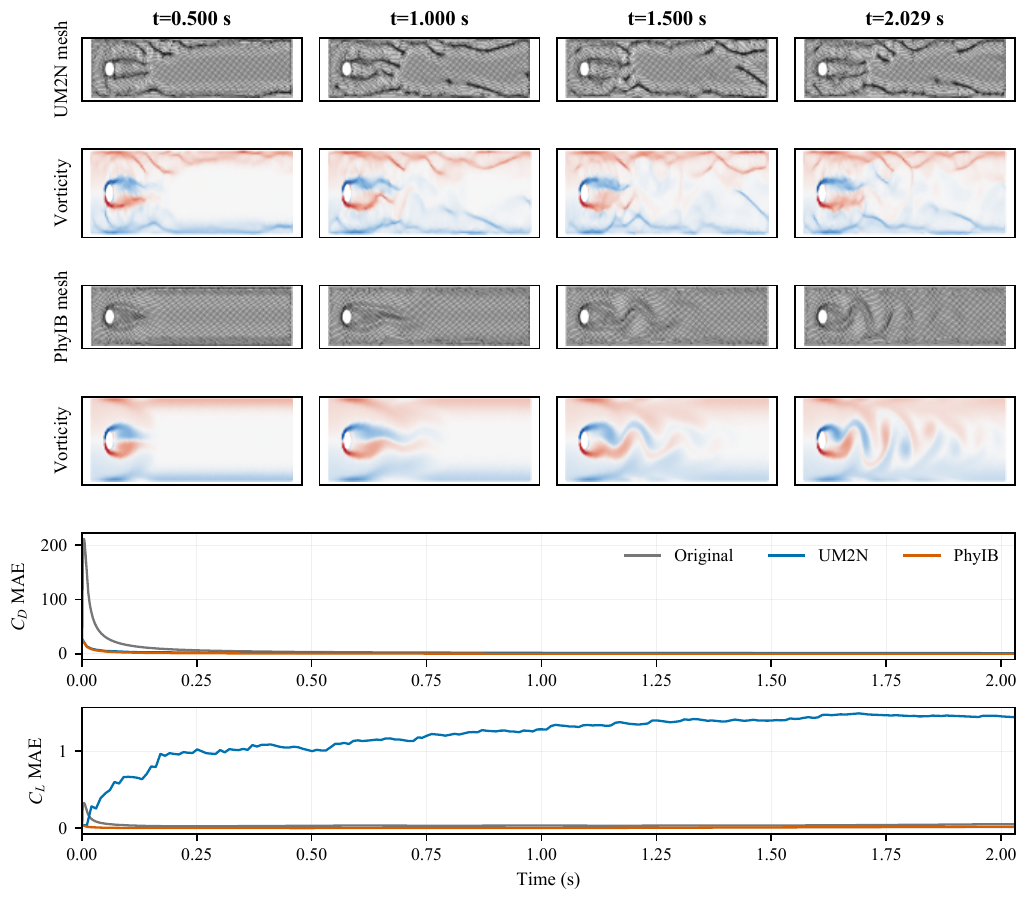}
\caption{Cylinder rollout: mesh/vorticity snapshots and cumulative mean absolute errors of drag and lift coefficients.}
\label{fig:app_cylinder_rollout}
\end{figure}

The rollout uses 2,030 steps with $\Delta t=0.001$, covering 2.03 s. Figure~\ref{fig:app_cylinder_rollout} shows snapshots at $t=0.5,1.0,1.5,2.029$ s: PhyIB concentrates nodes behind the cylinder and resolves the alternating wake vortices, whereas the UM2N vorticity field develops irregular structures. The lower panels show cumulative mean absolute errors of drag and lift coefficients $C_D$ and $C_L$; PhyIB maintains lower lift error than UM2N over the rollout.

\begin{figure}[H]
\centering
\includegraphics[width=.80\textwidth,height=.86\textheight,keepaspectratio]{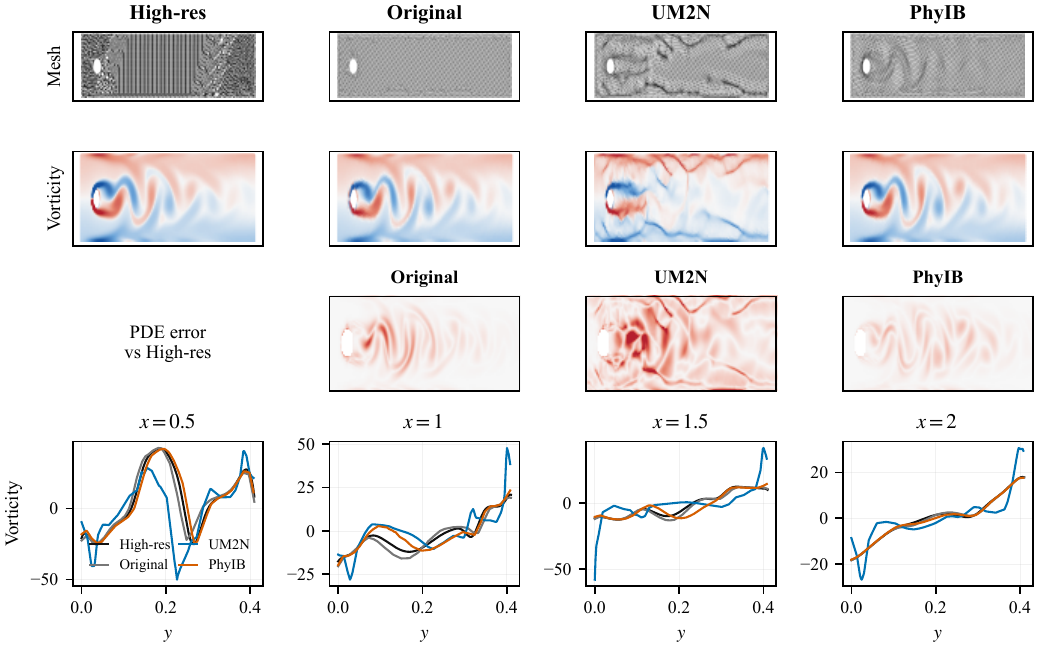}
\caption{Cylinder wake: meshes, vorticity, and absolute errors relative to the high-resolution reference. Bottom panels plot vorticity versus $y$ at fixed $x$.}
\label{fig:app_cylinder_wake}
\end{figure}

At $t=2.029$ s, Figure~\ref{fig:app_cylinder_wake} compares the wake with a reference mesh of 260,340 vertices and 517,912 triangles. The four lower panels plot vorticity along the transverse coordinate $y$ at streamwise positions $x\in\{0.5,1.0,1.5,2.0\}$. The PhyIB curves follow the reference peaks and troughs more closely than UM2N, which shows large deviations near the walls. The mean profile RMSE is 2.668 for Original, 10.789 for UM2N, and 3.071 for PhyIB: PhyIB improves over UM2N, while Original remains slightly more accurate on this diagnostic.


\begin{figure}[H]
\centering
\includegraphics[width=.80\textwidth,height=.86\textheight,keepaspectratio]{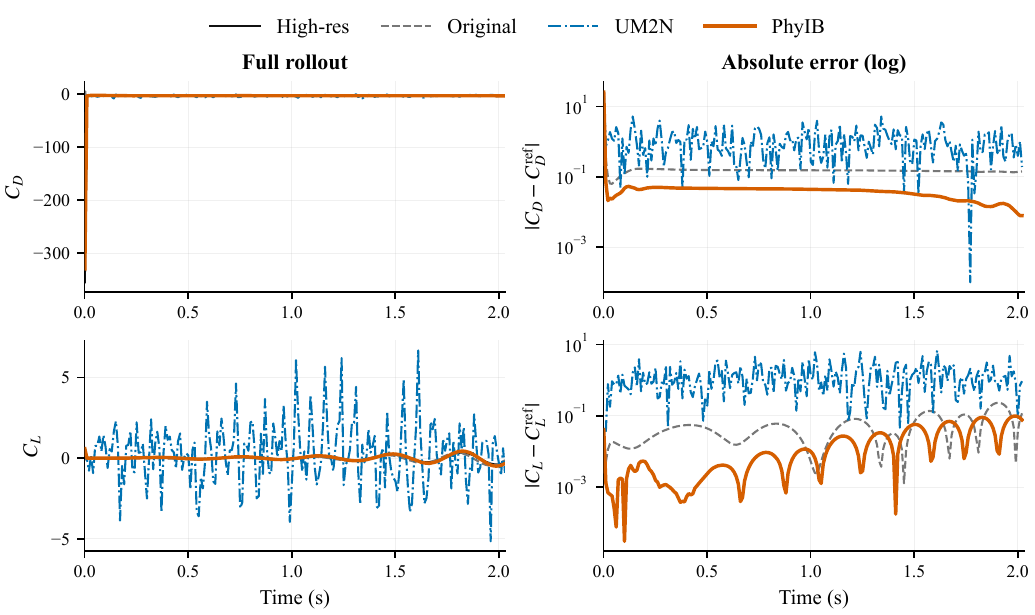}
\caption{Cylinder drag (top) and lift (bottom) coefficients (left), with absolute errors relative to High-res at reference timestamps on logarithmic axes (right).}
\label{fig:app_cylinder_forces}
\end{figure}

As shown in Figure~\ref{fig:app_cylinder_forces}, the orange PhyIB drag and lift curves closely follow the black high-resolution reference, while UM2N exhibits large lift oscillations. The absolute-error curves on the right confirm smaller force errors than UM2N over most of the rollout. This agreement indicates that the adapted mesh preserves the integrated force response as well as the wake structure.

\clearpage
\subsection{Near-wall Flow Results}
\label{app:boundary_layer}
We evaluate whether adaptation resolves the rapid near-wall velocity change in parallel-plate flow by comparing profiles at two downstream locations with a high-resolution reference.

The parallel-plate flow follows the UM2N example~\cite{zhang2024um2n} on $(0,2.2)\times(0,0.41)$:
\begin{equation}
\partial_t\mathbf v+(\mathbf v\cdot\nabla)\mathbf v-\nu\Delta\mathbf v+\nabla p=\mathbf0,
\qquad \nabla\cdot\mathbf v=0.
\label{eq:app_nearwall_pde}
\end{equation}
We set $\nu=0.001$, start from rest, and impose zero outlet pressure. We use CG2 velocity and CG1 pressure elements with incremental pressure correction.

This supplementary case uses no-slip upper and lower walls, inlet velocity $(1.5,0)$, time step $\Delta t=0.001$, and a 1,000-step rollout. The coarse and high-resolution reference meshes have 4,915 and 42,388 vertices, respectively. The shared saved state at step 990 is sampled along $y\in[0,0.1]$ at streamwise coordinates $x=0.1$ and $x=0.2$. We compute the velocity-profile RMSE against the high-resolution reference and report its percentage reduction relative to Original at each location.

\begin{figure}[H]
\centering
\includegraphics[width=.72\textwidth]{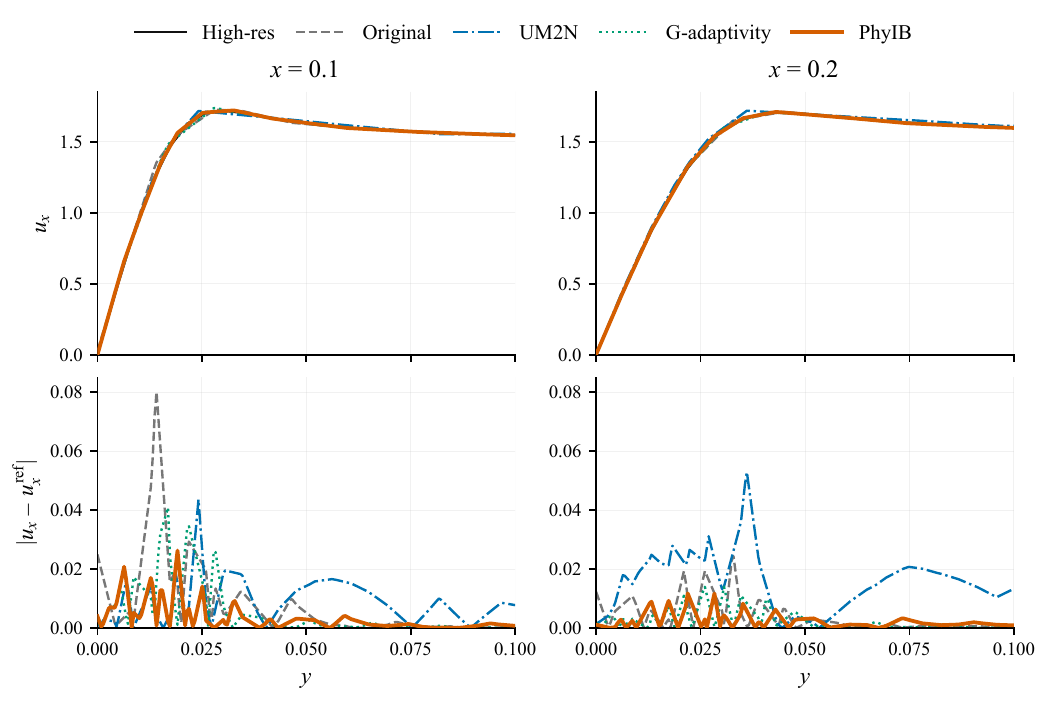}
\caption{Boundary-layer velocity profiles (top) and pointwise absolute errors relative to High-res (bottom) at two fixed $x$ locations.}
\label{fig:app_boundary_profiles}
\end{figure}

As shown in Figure~\ref{fig:app_boundary_profiles}, the orange PhyIB velocity profiles nearly overlap the black reference curves at both locations. The lower panels reveal smaller deviations in the steep near-wall region than Original and UM2N. Table~\ref{tab:app_boundary_profiles} quantifies this agreement: PhyIB reduces profile RMSE by $64.20\%$ and $47.27\%$, exceeding G-adaptivity at both locations; UM2N instead increases error at $x=0.2$. Thus, the adapted mesh better resolves the near-wall velocity distribution.

Original, UM2N, G-adaptivity and PhyIB complete the rollout. MA fails to converge within 1,000 mover iterations; M2N has 12 inverted elements at step 2; UGM2N has one at step 0; DMM violates the fixed boundary by $3.006\times10^{-2}$ at step 0. Table~\ref{tab:app_boundary_profiles} retains these failures.

\begin{table}[H]
\centering\small
\caption{Velocity-profile RMSE reduction (\%) at two streamwise locations. Status indicates rollout completion.}
\label{tab:app_boundary_profiles}
\begin{tabular}{lccc}
\toprule
Method & Profile ER at $x=0.1$ (\%) & Profile ER at $x=0.2$ (\%) & Status\\
\midrule
Original & 0.00 & 0.00 & Complete\\
MA & -- & -- & \textit{Fail}\\
M2N & -- & -- & \textit{Fail}\\
UM2N & 26.88 & -196.99 & Complete\\
UGM2N & -- & -- & \textit{Fail}\\
G-adaptivity & 34.19 & 36.26 & Complete\\
DMM & -- & -- & \textit{Fail}\\
PhyIB & 64.20 & 47.27 & Complete\\
\bottomrule
\end{tabular}
\end{table}

\Needspace{0.42\textheight}
\subsection{Computational Complexity}
\label{app:complexity}

We profile seed-666 checkpoints on an RTX 3090 with FP32 and batch size one. Helmholtz case 8 (513 nodes, 2,912 directed edges) and Circular case 00 (1,089 nodes, 6,272 directed edges) are used as matched probes. Learned methods use 20 warm-ups and 120 CUDA-synchronized repetitions. M2N, UM2N, and PhyIB-Mesh use resident neural forward inference with their native input semantics; UGM2N, G-adaptivity, and DMM use their native prediction routines on preloaded samples, excluding disk I/O and PDE solution. For the classical MA baseline, we use the same preloaded monitor on each probe and time only \texttt{MongeAmpereMover.move()} over 3 warm-ups and 30 repetitions, excluding mesh construction, disk I/O, and the underlying PDE solve.

\begin{table}[H]
\caption{Active parameter counts and inference/adaptation latency (ms); $^\dagger$ marks an invalid timed Circular output.}
\label{tab:complexity}
\centering
\small
\setlength{\tabcolsep}{4pt}
\begin{tabular}{lccc}
\toprule
Model & Active parameters & Helmholtz (ms) & Circular (ms) \\
\midrule
MA & N/A & $433.725\pm67.334$ & $623.934\pm57.356$ \\
M2N & $1{,}854{,}483$ & $5.760\pm0.942$ & $7.923\pm0.738$ \\
UM2N & $1{,}839{,}103$ & $5.498\pm0.212$ & $7.607\pm0.407$ \\
UGM2N$^\dagger$ & $36{,}374{,}786$ & $726.347\pm30.424$ & $2142.676\pm68.845$ \\
G-adaptivity & $1{,}839{,}103$ & $48.475\pm0.713$ & $48.406\pm1.381$ \\
DMM & $1{,}100{,}882$ & $10.237\pm0.275$ & $12.442\pm0.609$ \\
PhyIB-Mesh (Full) & $2{,}907{,}751$ & $14.305\pm0.461$ & $14.535\pm0.404$ \\
\bottomrule
\end{tabular}
\end{table}
As reported in Table~\ref{tab:complexity}, PhyIB-Mesh uses 2.91M active parameters and takes approximately 14.3--14.5 ms on the two probes. Its latency is higher than M2N and UM2N, but lower than G-adaptivity and UGM2N. Relative to classical MA, the mover-only speedups are approximately $30.3\times$ on Helmholtz and $42.9\times$ on Circular.

\Needspace{0.48\textheight}
\subsection{Sensitivity to the Sparse Prior}
\label{app:prior_sensitivity}
\begin{figure}[H]
\centering
\includegraphics[width=.98\textwidth]{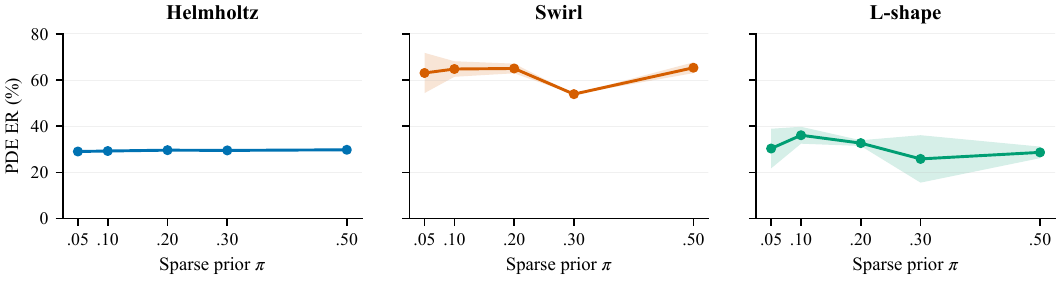}
\caption{Sparse-prior sensitivity on UM2N. The horizontal axis shows prior $\pi$ and the vertical axis shows PDE ER (\%). Lines denote means and shaded bands show $\pm1$ sample SD across three training seeds.}
\label{fig:app_pi_prior_sensitivity}
\end{figure}

\noindent\begin{minipage}[t]{0.48\linewidth}
\raggedright
We vary the Bernoulli prior $\pi\in\{0.05,0.10,0.20,0.30,0.50\}$ using UM2N, three training seeds, and fixed evaluation sets, with other settings unchanged. As shown in Figure~\ref{fig:app_pi_prior_sensitivity} and Table~\ref{tab:pi_prior_sensitivity}, Helmholtz remains nearly flat at $29.04$--$29.78\%$ ER. Swirl drops to $53.91\%$ at $\pi=0.30$, while L-shape reaches its highest mean at $\pi=0.10$ and has its largest seed variation at $\pi=0.30$. These differing trends indicate task-dependent sensitivity. These results indicate that $\pi=0.10$ is not a special setting required to obtain good PDE performance.
\end{minipage}\hfill
\begin{minipage}[t]{0.48\linewidth}
\centering
\small
\captionof{table}{Sparse-prior sensitivity on UM2N: PDE ER (\%, mean $\pm$ sample SD across three training seeds).}
\label{tab:pi_prior_sensitivity}
\setlength{\tabcolsep}{3pt}
\begin{tabular}{cccc}
\toprule
$\pi$ & Helmholtz & Swirl & L-shape \\
\midrule
0.05 & $29.04 \pm 0.48$ & $63.08 \pm 8.72$ & $30.32 \pm 8.58$ \\
0.10 & $29.24 \pm 0.36$ & $64.80 \pm 3.37$ & $36.09 \pm 3.67$ \\
0.20 & $29.62 \pm 0.26$ & $65.04 \pm 2.09$ & $32.67 \pm 1.24$ \\
0.30 & $29.51 \pm 0.10$ & $53.91 \pm 0.08$ & $25.82 \pm 10.28$ \\
0.50 & $29.78 \pm 0.55$ & $65.35 \pm 2.33$ & $28.66 \pm 2.51$ \\
\bottomrule
\end{tabular}
\end{minipage}

\clearpage
\end{document}